\documentclass{article}
\PassOptionsToPackage{numbers,sort,compress}{natbib}
\usepackage{arxiv}
\usepackage[multidot]{grffile}
\usepackage[utf8]{inputenc} 
\usepackage[T1]{fontenc}    
\usepackage{hyperref}       
\usepackage{url}            
\usepackage{booktabs}
\usepackage{amsfonts}
\usepackage{caption}
\usepackage{subcaption}
\usepackage{nicefrac}       
\usepackage{microtype}      
\usepackage{natbib}
\usepackage[table,xcdraw]{xcolor}
\usepackage{microtype}
\usepackage{pgfplots}
\usetikzlibrary{patterns}
\pgfdeclarelayer{bg}    % declare background layer
\pgfsetlayers{bg,main}  % set the order of the layers (main is the standard layer)
\usepackage{multirow}
\pgfplotsset{compat=newest}% use newest version
\usepackage[title]{appendix}
\usepackage{float}
\usepackage{amsmath,bm,mathtools,mathrsfs} 
\usepackage{amssymb}
\usepackage{amsthm}
\usepackage{xspace}
\usepackage{enumerate}
\usepackage[ruled,noend]{algorithm2e}
\usepackage[table,xcdraw]{xcolor}

\definecolor{darkblue}{rgb}{0.0,0.0,0.55}
\hypersetup{
  colorlinks = true,
  citecolor  = darkblue,
  linkcolor  = darkblue,
  citecolor  = darkblue,
  filecolor  = darkblue,
  urlcolor   = darkblue,
}

\pgfdeclarepatternformonly{my crosshatch dots}{\pgfqpoint{-1pt}{-1pt}}{\pgfqpoint{3pt}{3pt}}{\pgfqpoint{4pt}{4pt}}%
{
    \pgfpathcircle{\pgfqpoint{0pt}{0pt}}{.5pt}
    \pgfpathcircle{\pgfqpoint{2pt}{2pt}}{.5pt}
    \pgfusepath{fill}
}

\numberwithin{equation}{section}

\newcommand{\norm}[1]{\|{#1}\|}
\newcommand{\bnorm}[1]{\left\lVert#1\right\rVert}

\numberwithin{equation}{section}

\newcommand{\argmin}{\operatornamewithlimits{argmin}}
\newcommand{\argmax}{\operatornamewithlimits{argmax}}
\begin{document} 
\tikzset{mark options={mark size=5, line width=3pt}}
%\title{A Novel Approach for Training Deep Radial Basis Function Networks Inspired by Metric Learning for the Nearest Centroid Classifiers}
\title{ Deep-RBF Networks Revisited: \\ Robust Classification with Rejection}
\author{\name Pourya Habib Zadeh \email {poorya.habibzadeh@gmail.com}\\ 
\name Reshad Hosseini \email {reshad.hosseini@ut.ac.ir}\\
\addr{University of Tehran} \\
\name Suvrit Sra \email {suvrit@mit.edu}\\
\addr{Massachusetts Institute of Technology}}
\maketitle

\vspace{2.5ex}
\begin{abstract}
%Common deep neural networks have a non-compact decision surface after training. This property cause several drawbacks of these networks, one of which is the vulnerability to adversarial attacks.
One of the main drawbacks of deep neural networks, like many other classifiers, is their vulnerability to adversarial attacks. An important reason for their vulnerability is assigning high confidence to regions with few or even no feature points. By feature points, we mean a nonlinear transformation of the input space extracting a meaningful representation of the input data. On the other hand, deep-RBF networks assign high confidence only to the regions containing enough feature points, but they have been discounted due to the widely-held belief that they have the vanishing gradient problem. In this paper, we revisit the deep-RBF networks by first giving a general formulation for them, and then proposing a family of cost functions thereof inspired by metric learning. In the proposed deep-RBF learning algorithm, the vanishing gradient problem does not occur. We make these networks robust to adversarial attack by adding the reject option to their output layer. Through several experiments on the MNIST dataset, we demonstrate that our proposed method not only achieves significant classification accuracy but is also very resistant to various adversarial attacks.

\end{abstract}

%%%%%%%%%%%%%%%%%%%%%%%%%
\section{Introduction}

Deep Neural Networks are stacks of several layers of simple computational units which can collectively compute highly complex and structured functions~\cite{montufar2014number}.
They have recently outperformed other machine learning methods in various tasks, such as image classification~\cite{he2016deep,szegedy2017inception,zoph2017learning}, object detection~\cite{ren2015faster,redmon2018yolov3,he2017mask}, and speech recognition~\cite{,chiu2018state,bahdanau2016end}. However, it has been shown that common deep learning methods, like many other machine learning algorithms for classification, can be easily fooled by a small imperceptible change in the input~\cite{szegedy2013intriguing, nguyen2015deep}. This vulnerability is due to a flaw in their generalization mechanism, that is, some feature points in the support of the feature space distribution get low confidence while feature points in the areas far from the support obtain high confidence.

One reason for such behavior could lie in the linear classifier they use at their last layer. Almost all modern deep neural networks used for classification are made up of a feature extractor and a linear classifier. Ideally, the goal of the feature extractor is to map data to a feature space where different classes are linearly separable. Although the linear classifiers are very efficient in classifying the linearly separable classes, they force the model to assign high confidence to regions which are far from the decision boundary, i.e., regions without any data-point.
An adversarial attack can easily make small changes in many dimensions of the input image to land in these areas of the feature space and lead the network to misclassify the perturbed example with high confidence. According to the linear explanation of adversarial examples~\cite{goodfellow2014explaining}, the attack only needs to find the right direction of perturbation to reach those areas.
%An adversarial attack can then easily change the pixels of 
%These wide regions of the feature space in which no data-point exist correspond to vast regions in the input space, not necessarily belong to a specific category.

Unlike common networks, RBF networks do not use linear classifiers, and instead, they have {\it Radial Basis Function} units in their last layers. \citet{goodfellow2014explaining} claimed that RBF networks are naturally immune to adversarial and rubbish-class examples, in the sense that they give low confidence to such examples. RBF units, unlike linear units, are activated within a well-circumscribed region of their input-space \citep{lecun1998gradient}. In this case, the aim of the feature extractor network would be mapping the data to a new representation where the data for each category form a cluster. By adding a reject option in our classifier, e.g., whenever we are far from the center of the clusters, we can make the regions of each class in the feature space finite and narrow. \citet{bottou1992local} called this property of RBF networks, as a local classifier, the ability of {\it rejection by lack of data}. 
 We hypothesize that this ability of RBF networks make them not only robust to adversarial examples but a perfect choice in tasks where dealing with {\it negative examples}\footnote{ A negative example or a null example belongs to none of the classes or categories.} is necessary. This claim follows from the assumption that non-typical patterns usually fall outside those clusters.

%A network with {\it Radial Basis Function} (RBF) units may overcome this problem. \citet{goodfellow2014explaining} claimed that RBF networks are resistant to adversarial and rubbish-class examples. Also, RBF units in the last layer of a deep network, unlike softmax, are activated within a well-circumscribed region of the network's input-space \citep{lecun1998gradient}. Because non-typical patterns usually fall outside of that region, RBF networks are more appropriate for rejecting {\it negative examples}\footnote{ A negative example or a null example belongs to none of the other classes or categories.}, whereas in ordinary CNNs, assigning high confidence to a region without data-point is a common phenomenon. \citet{bottou1992local} called this property of RBF networks (as a local classifier) the ability of {\it rejection by lack of data}.

The first deep-RBF network, called \textbf{LeNet5}, is introduced in the seminal work of~\citet{lecun1998gradient}. They proposed a simple form of the deep-RBF networks and two cost functions for learning, in addition to performing many tricks for avoiding the vanishing gradient problem. On MNIST, they  achieved significant results for their method. After LeNet5, \citet{simard2003best} came with a slightly different model with a softmax output layer instead of the RBF output layer and used the cross-entropy loss. They reported that these modifications yield a small outperformance in the accuracy and a faster optimization procedure. To the best of our knowledge, the RBF output layer, surprisingly, has disappeared from the literature of deep learning afterward, and the linear classifier output layer has become the standard choice for the output layer of deep networks.

In this paper, we reintroduce the deep-RBF networks with a formulation more general than LeNet5. We also mention a simple yet efficient way to incorporate the reject option in these networks. Inspiring from the literature on metric learning, we propose an example of a class of cost functions, in the optimization of which the vanishing gradient problem does not occur. We then present the soft version of the proposed cost and its probabilistic viewpoint.  Using a variety of experiments on MNIST, we demonstrate that this method can be trained efficiently to achieve high accuracy in the task of classification. 

Furthermore, we show that by assigning a threshold for rejection and incurring a small cost on the classification, we can make the deep-RBF networks resistant to multiple adversarial attacks. We show that this robustness can be further improved using noise injection in training and decreasing the threshold of rejection. Finally, we examine the cross-model generalization of the adversarial examples between our methods and the ordinary CNN, together with the effect of noise injection on the percentage of transferrable adversarial examples. The results show, unlike other deep neural networks, there is a relatively low percentage of adversarial examples transferrable between deep-RBF networks and CNNs.

%In the section ``TO-DO list'', there is a list of some important works we believe we should do to advance this research. At the end of the paper, the section ``Future works'' proposes the unsupervised version of the method, in addition to introducing another loss function for training the deep-RBF network. These methods are so crude that we haven't empirically examined them yet.

%%%%%%%%%%%%%%%%%%%%%%%%%
\section{Deep-RBF networks}
\label{sec:defdeep-RBF}

A radial basis function is a real-valued function which depends only on the distance from the origin or a point~\cite{buhmann2003radial}. In this paper, we use the following formula with $\ell_p$-norm
% for $p > 1
to define the RBF unit:
\begin{equation}
\phi(x) = \bnorm{A^{T}x+b}_p^p,
\end{equation}
in which $x\in \mathbb{R}^{n}$, $A \in \mathbb{R}^{n\times l}$, and $b \in \mathbb{R}^{l}$ for $l \leq n$. If $p=2$, it is trivial to see that this RBF unit is equivalent to $(x-\mu)^T \Sigma (x-\mu)$, given the factorizations $\Sigma:=A A^T$ and $b:=-A^T\mu$.

\begin{figure}[ht]
\centering
        \includegraphics[totalheight=6.7cm]{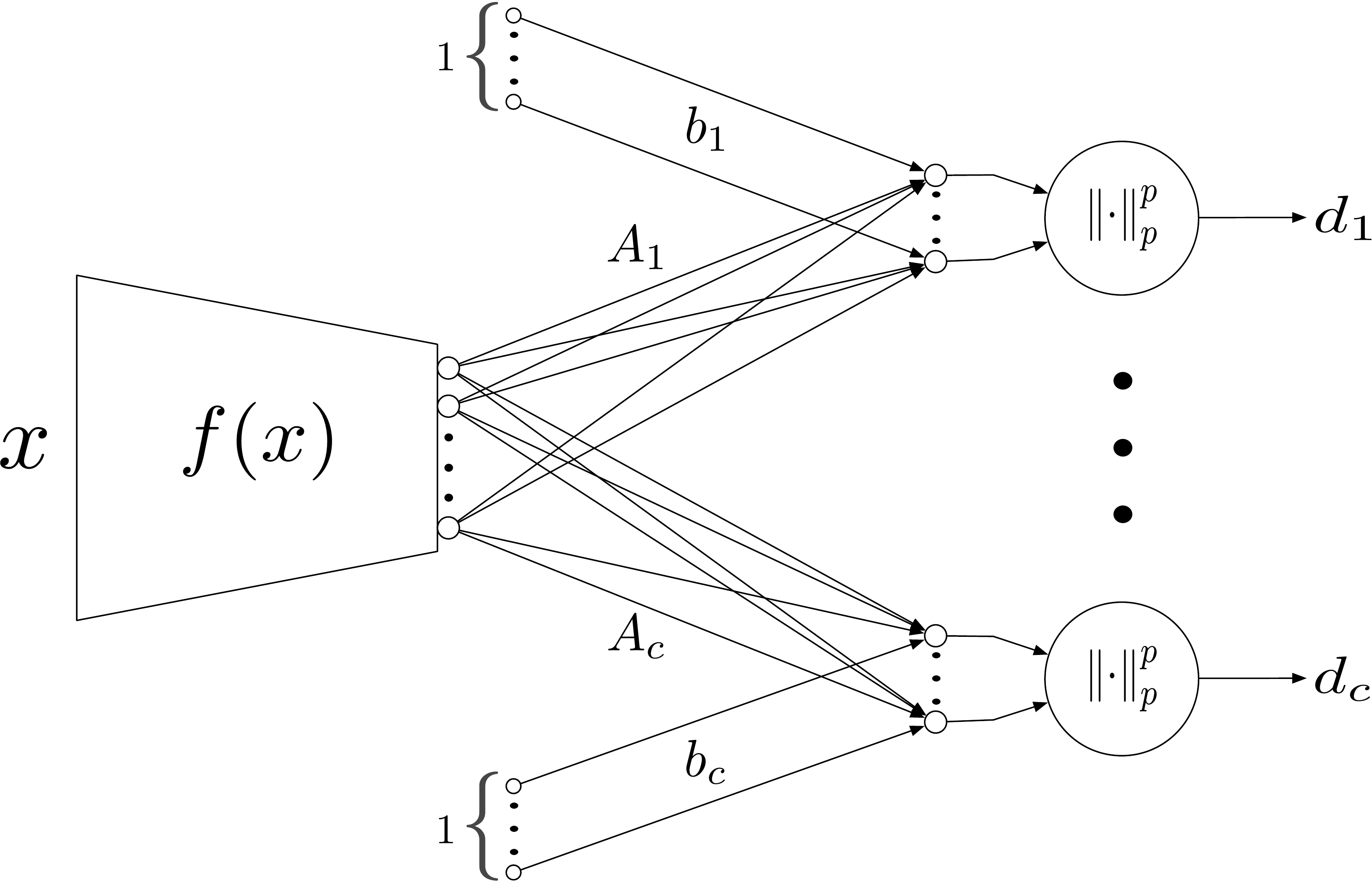}
    \caption{{\small The diagram of a deep-RBF network with $c$ classes. As the diagram shows, there is an RBF unit for every class on the transformation of the input, i.e., $f(x)$.}}
    \label{fig:diag1}
\end{figure}

In the \emph{shallow-RBF network}, there exists a unique RBF unit for each class, and the output of the network is
\begin{equation}
d_k(x) = \bnorm{A_k^Tx+b_k}_p^p,
\end{equation}
 for $k \in \{1, 2, \dots, c\}$. In the evaluation phase, the category with the smallest distance would be selected as the correct class. In the \emph{deep-RBF network}, as its name suggests, a function, say $f(\cdot)$, transforms the raw input data $x$ to high-level features, then they enter the RBF units:
\begin{equation}
d_k(x) = \bnorm{A_{k}^{T}f(x)+b_k}_p^p, \qquad k\in\{1, \dots, c \}.
\label{eq:distance}
\end{equation}
The above equation is the definition of deep-RBF networks, about which we talk in the rest of this paper. Note that the outputs of the network are not considered to be the probability of classes; an output layer (like softmax) may turn them into probabilities. The schematic diagram of this network appears in Fig.~\ref{fig:diag1}.

The decision rule of deep-RBF network, in the evaluation, is as follows:
\begin{equation}
\hat{y}(x) = \argmin_{k \in \{1, \dots, c \}}d_k(x).
\label{eq:NCC}
\end{equation}
One can look at this method as a generalization of the {\it nearest centroid classifiers}. In these classifiers, each class has a centroid, and the category of the nearest one would be selected~\citep{tibshirani2002diagnosis}. As an example, Gaussian discriminant analysis with uniform prior is the nearest centroid classifier with the metric $d_k(x, \mu_k) = (x- \mu_k)^T\Sigma_k^{-1}(x- \mu_k)$, where $\mu_k$s are the per-class centroids~\citep{Murphy2013Machine}. To connect the nearest centroid classifier with the deep-RBF network, assume Eq.~\ref{eq:distance} is the parametric distance of the nearest centroid classifier. 
%Therefore, we can consider deep-RBF networks as \textit{non-linear metrics  for nearest centroid classifiers}.

\subsection{Deep-RBF networks in the literature of deep learning}

\begin{figure}[ht]
\centering
        \includegraphics[totalheight=4.3cm]{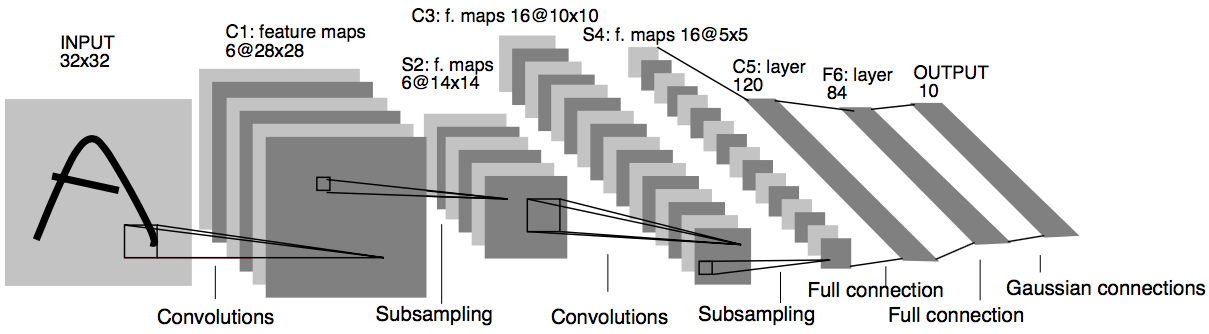}
    \caption{{\small Architecture of LeNet5. Source: Fig. 2 from \citep{lecun1998gradient}.}}
    \label{fig:LeNet5}
\end{figure}

As far as we know, LeNet5~\cite{lecun1998gradient} is the only form of the deep-RBF networks introduced in the literature of deep learning. As shown in Fig.~\ref{fig:LeNet5} this model has some convolutional, subsampling and fully connected layers. These layers are to compute the {\it feature vector} of the input image, which is associated with $f(x)$ in the deep-RBF network. At the end of its last fully connected layer (whose activation function is a scaled-tanh), there is a layer with so-called Gaussian connections. In this layer, there exists the {\it parameter vector} $W_k$ for each class (equivalent to the negative bias in the deep-RBF network), and each output is the Euclidean distance between the feature vector and the corresponding parameter vector. From deep-RBF networks viewpoint, all covariances are identity matrices in this network, and the output distance is $\hat{d}_k(x) = \norm{f(x)-W_k}_2^2$. In the evaluation phase, LeNet5 selects the output with the smallest value, whose parameter vector is nearest to the feature vector.

% The architecture of this network from the point of view of deep-RBF networks is drawn in Fig.~\ref{fig:diag2}.

%\begin{figure}[h]
%\centering
%        \includegraphics[totalheight=6.2cm]{figures/diag2.pdf}
%    \caption{{\small The schematic diagram of LeNet5. The function $f(x)$ consists of some convolutional, subsampling and fully connected layers. All covariances are identity matrices.}}
%    \label{fig:diag2}
%\end{figure}

For training LeNet5, the authors proposed two strategies with two different loss functions. In the first strategy, parameter vectors are fixed. The authors chose the binary image of a character's ASCII code (binary with components $+1$ and $-1$) as the value of its fixed parameter vector\footnote{ They claimed this choice of parameter vector works better than one-hot coding and random binary vectors.}. Another trick they used to prevent the gradient vanishing problem (resulting from the saturation of sigmoids) is choosing the scale parameter of the scaled-tanh activation functions of the last fully connected layer to be $1.7159$. By doing so, the components of the parameter vectors, $+1$ and $-1$, are the points of sigmoids with maximum curvature. The following mean squared error function is the loss function of the first strategy:
\begin{equation}
J = \frac{1}{N} \sum_{i=1}^{N}\hat{d}_{y_i}(x^{(i)}) = \frac{1}{N} \sum_{i=1}^{N}\norm{f(x^{(i)})-W_{y_i}}_2^2,
\label{eq:LeNet5V1}
\end{equation}
where $y_i$ is the correct class of input data $x^{(i)}$ and $N$ is the number of data-points. In the second strategy, they also trained the parameter vectors. They are initialized with the images of their corresponding characters' ASCII code, just like the previous strategy. The authors also used the other tricks to have an efficient optimization. The loss function of this strategy is
\begin{equation}
J_{\text{LeNet5}} = \frac{1}{N} \sum_{i=1}^{N}\bigg(\hat{d}_{y_i}(x^{(i)}) + \log\Big(\exp(-\lambda)+\sum_{j}\exp\big(-\hat{d}_j(x^{(i)})\big)\Big)\bigg),
\label{eq:LeNet5V2}
\end{equation}
where $\lambda > 0$. The second, new, term in this loss function aims to prevent collapsing, in which all the parameter vectors are equal. The constant $\lambda$ prevents the penalties of classes which are already large enough from being pushed further up. As they pointed out, this penalty term acts as a prior for the model -- that the input image can come from only one of the classes or a negative class.

As we mentioned in the previous section, the output layer of LeNet5 and these cost functions were substituted by the softmax output layer and cross-entropy loss function afterward. One of the main reasons for this substitution lies in the fact that this new setting relieves us of doing many tricks to prevent the gradient vanishing problem. Besides, these tricks were tuned for MNIST, which makes it very difficult to reuse the network for other datasets.

In this paper, we generalize and modify LeNet5, with its second training strategy, from three perspectives. First, generalizing the distance of LeNet5; second, throwing all the saturating units away in order to have an efficient training without multiple tricks; third, proposing other loss functions which outperform that of LeNet5 in terms of classification accuracy and the ability of rejection by lack of data, thereby resisting to the adversarial examples better. The first two generalizations can be achieved by substituting LeNet5 with the deep-RBF network and using other activation functions instead of the sigmoid. The third generalization is the subject of the next section.

%(also switching from the above MSE-like cost functions to other cost functions to nullify the bad effect of the saturating units in the last layer)

%%%%%%%%%%%%%%%%%%%%%%%%%
\section{Learning algorithm}
\label{sec:learning}

According to the decision rule of the deep-RBF networks, the primary goal of training such networks is to find distances, or outputs, that are small for their associated class and big otherwise. Conceptually, this is a task similar to metric learning problems although the original metric learning problem is weakly-supervised, with similar and dissimilar pairs. We can substitute the distance of pairs of similar objects with the distance of the correct class and the distance of pairs of dissimilar objects with the distance of the wrong classes. Inspiring from the metric learning methods, we propose the following cost function:
\begin{equation}
J_{\text{ML}} = \sum_{i=1}^{N}\bigg(d_{y_i}\big(x^{(i)}\big) + \sum_{j \notin y_i}\max{\Big(0, \lambda-d_j\big(x^{(i)}\big)\Big)} \bigg),
\label{eq:J1}
\end{equation}
where $\lambda > 0$ and $y_i$ is the correct class of the input $x^{(i)}$. We name it {\it Metric Learning} (ML) loss. While the first term in this cost function, like in LeNet5, aims to decrease the distance of the correct class~\footnote{ Or as a nearest centroid classifier, it decreases the distance between the $i$-th sample and its corresponding centroid.}, the other term, using the hinge loss, pushes the distance of the wrong classes outside a margin and also prevents collapsing. Note that this cost function is just an example of numerous cost functions available in the metric learning literature~\citep{bellet2013survey}.  We can propose many alternative costs, each of which has its merits and demerits.

In theory, the value of parameter $\lambda$ should not affect the classification accuracy.
Let us assume $\lambda^{\prime} = \alpha \lambda$, where $\alpha$ is a positive real number. The distances $d_{k}$s, given in~\eqref{eq:distance}, depends on $A_k$s, $b_k$s and the parameters of the function $f(\cdot)$. Using the notation $J_{ML}(\{A_k, b_k\}_{k=1}^{c}, f(\cdot) ;\lambda)$ for showing the parameters in the $J_{ML}$ cost function, it is trivial to see that
\begin{equation*}
J_{\text{ML}}(\{\alpha A_k, \alpha b_k\}_{k=1}^{c}, f(\cdot) ;\lambda^{\prime}) = \alpha J_{\text{ML}}(\{A_k, b_k\}_{k=1}^{c}, f(\cdot) ;\lambda).
\end{equation*}
Therefore, the optimal distances for the ML loss with $\lambda^{\prime}$ and $\lambda$ only differ in a scale. Since the decision rule of the deep-RBF network is scale invariant with respect to distance, the classification accuracy of the two cases would be the same. However, due to the non-convexity of the loss, this parameter can affect the optimization procedure. We will see in Section~\ref{sec:Result} that in practice for a wide range of $\lambda$, the classification accuracy on the test dataset does not change significantly, so choosing a proper $\lambda$ is quite easy.

%{ \color{red} \# THE FOLLOWING DOES NOT HOLD
%The following theorem states that this cost function w.r.t the parameters of the last layer behaves well in the optimization procedure — just like a network with the softmax output unit.
%
%\begin{theorem}
%The cost function~\ref{eq:J1} with respect to the matrices $A_k$s and $b_k$s has exactly one local minimum in each orthant\footnote{ The hyper-octant which is the generalization of the quadrant.} (at least for $p=1$ or $p=2$). Also, the values of the cost function in all of these local minimums are equal; hence, each local minimum is a global minimum.
%\end{theorem}
%
%\begin{proof}
%NOTHING
%\end{proof}
%}

Taking it into account that the log-sum-exp function is a smooth approximation to the maximum function, one can notice a difference between ML loss and the loss function of LeNet5. In ML loss, all the wrong classes inside the margin incur a penalty, but in loss~\ref{eq:LeNet5V2}, just the one with the minimum distance inside the margin incurs the penalty. Moreover, if the smallest distance is the correct class and inside the margin, the two terms of LeNet5 cost for that data-point cancel out each other while ML always tries to decrease the distance of the correct class. We will see that these differences lead to an improvement in both accuracy and robustness of the network trained with the ML loss.

\subsection{Rejection}

After training, in the evaluation phase, deep-RBF's decision rule~\ref{eq:NCC} determines the category of the input among $c$ classes. However, as explained before, these networks can reject by lack of data as a result of using RBF units at the last layer of the network. In other words, we expect that the regions in the feature space without data get low confidence. Adding a reject option can be beneficial in many tasks where a negative class exists. 
%For example, the {\it background} class in object detection, associated with a region in which no object exists, or the {\it blank token} in automatic speech recognition, corresponding to the absence of any phonemes or characters in a frame of the input signal. 
Unlike the standard non-negative examples, a negative sample usually does not have a specific pattern, and the set of all negative samples, in a given task, is vast. This vast region of negative examples often makes it impossible to provide a training dataset spanning all over that region. Consequently, an ordinary network which treats the negative and non-negative examples, in the same way, may have an issue because the training samples for the class of negative examples are incomplete. Furthermore, as we observe in the next section, the rejection will also make the network robust to adversarial attacks.

The easiest way to incorporate the reject option in this classifier is to compare the distance of the selected class with a threshold using the condition $\min_{i \in \{1, \dots, c\}}d_i(x) > T$. If the condition holds, then the input data will be rejected; i.e., it belongs to none of the classes and gets a negative class label. We can imagine several other ways to incorporate the rejection; for example, we can assign different thresholds for different classes. In this case, if the minimum distance belongs to class $k^*$, then $T_{k^*}$ will be the threshold for rejection. However, as reported in Section~\ref{sec:Result}, this simple choice works well empirically.

While not common, there is the possibility that some regions in the input space which have to be rejected wrongly map to the non-rejected parts in the feature space. The deep-RBF network can even learn where is the negative area in the input space if a set of negative examples is available in the training phase too. Similar to the ML loss, we can use the following cost for these examples:
\begin{equation} 
\label{eq:negLoss}
 \sum_{j}\max{\Big(0, \lambda-d_j\big(x^{(i)}\big)\Big)},
\end{equation}  
where $x^{(i)}$ is a negative sample. This cost aims to increase the distance of all existing categories, up to a margin. Hence, for a given threshold, we will make the non-rejected areas in the input space more compact, improving the ability of rejection by lack of data.

\begin{figure}[t]
\centering
        \includegraphics[clip, trim=0cm 3cm 0cm 2.5cm, totalheight=5.6cm]{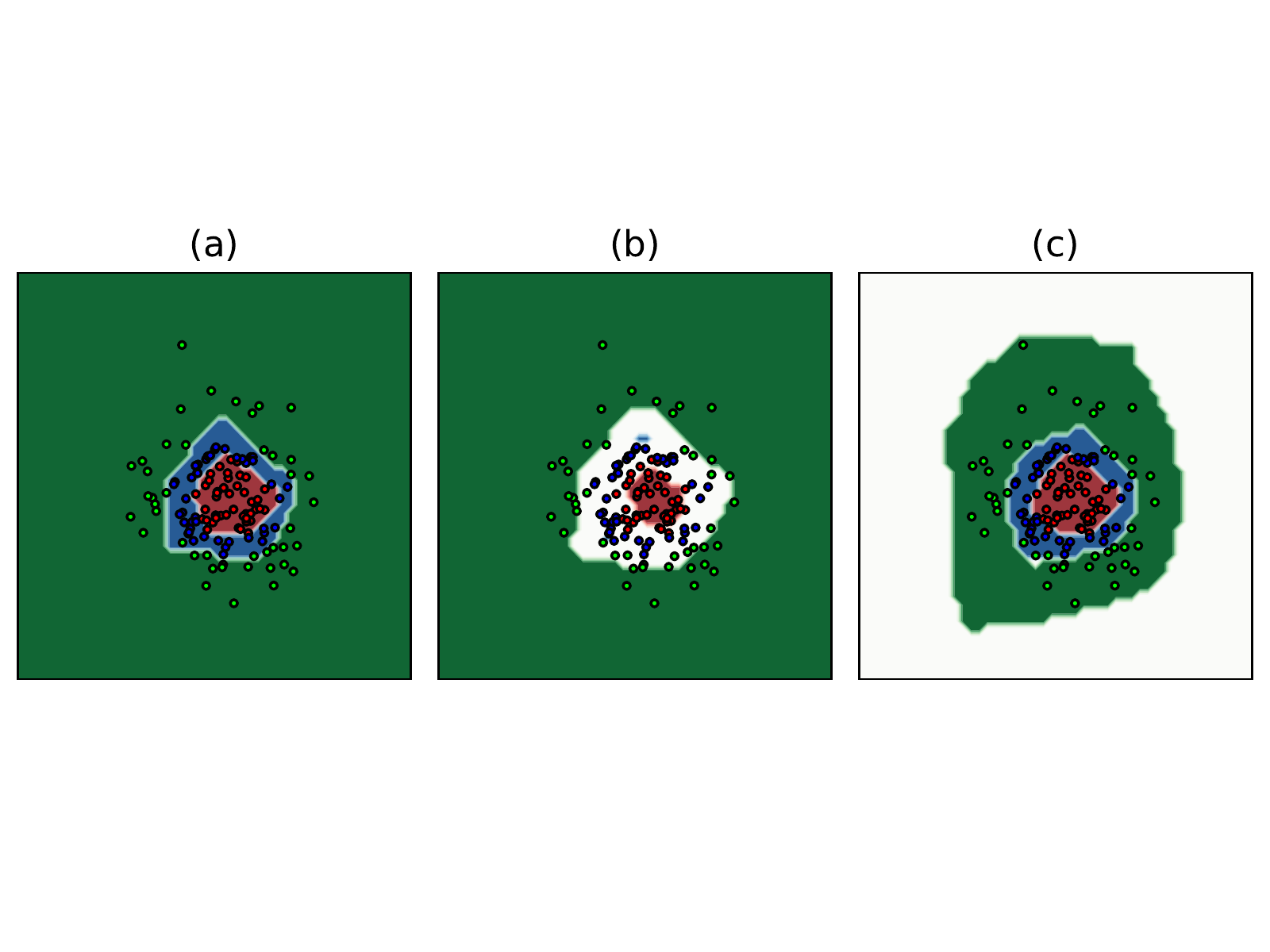}
 \caption{{\small The decision surfaces of two trained networks on a 2D toy dataset. Different classes correspond to different colors, and the rejected sections are colorless. (a) The left plot is the decision surface of a simple one hidden layer MLP with the softmax output layer. (b) The plot in the middle is the same network but with a threshold on the confidence of the softmax. (c) The right plot is the decision surface of a deep-RBF network with reject option when the same one hidden layer MLP is the feature extractor. As the plot in the middle displays, the low confidence in the network with the softmax output layer corresponds only to high ambiguity, not the lack of data. The network with the RBF output layer, on the other hand, can assign low confidence to the regions without enough data-point or with high ambiguity.}}
\label{fig:toy1}
\end{figure}

In order to visually illustrate the ability of rejection by lack of data, which the deep-RBF networks have, we experiment with a 2D toy dataset. Fig.~\ref{fig:toy1} displays the results of the experiment. This figure shows the decision surfaces of two trained networks on a toy dataset with three classes. The left plot is the decision surface\footnote{ Here, the {\it decision surface} of one class refers to the regions associated with that class; it differs from {\it decision boundary}.} of a simple one hidden layer MLP with softmax output layer. The plot in the middle is the same network but with a threshold on the confidence of the output. If the most significant output of the softmax is lower than an arbitrary threshold (here, $0.9$), we reject the decision of the network, i.e., assign a negative class. The rejected sections are colorless. The right plot is the result of a deep-RBF network when the same one hidden layer MLP is used as the feature extractor network $f(x)$. We used $\ell_1$-norm for the distance of this deep-RBF network. Also, the dimensionality of the output is two.

It is crystal clear that the rejection made by the threshold on softmax confidence is not plausible at all because the regions without any data have high confidence. As the plot in the middle clearly shows, the low confidence in vanilla softmax networks is only associated with the high ambiguity, not the lack of data. The ambiguity is high whenever the classifier is unsure which category it has to select. However, our proposed method can assign low confidence, i.e., large distance, to the regions without enough data-point, besides rejection by the ambiguity. This is the crucial advantage of our proposed method over the classifiers with the softmax output layer. If we assign a threshold, the deep-RBF networks will reject vast regions without data, and not lose much classification accuracy at the same time.

\subsection{The soft version of the metric learning loss and its probabilistic viewpoint}
\label{sec:deepRBF}

It is common in the literature of deep learning to look at the classifiers from a probabilistic point of view. In other words, consider one output of the network as the probability of a category. For example, the softmax function after the last layer of neural networks transforms the outputs of the network to probabilities, but does not affect the decision of the network in the evaluation. In this setting, training the network is merely minimizing the cross-entropy between the training data and the model's predictions~\citep{goodfellow2016deep}. In order to make a connection between the learning of deep-RBF networks and maximum likelihood learning, we introduce the soft version of the ML loss. We show that this cost is equivalent to the {\it Negative Log-Likelihood} when choosing a specific output layer.

% by which the network's output would be the probabilities of each category.
% We can link the ML loss to  the by introducing the soft version of ML cost.
%In order to develop a probabilistic viewpoint of the SoftML cost function, we propose an output unit layer for the deep-RBF network. With this output unit, we can train the network efficiently using the maximum likelihood principle.

Using the log-sum-exp smooth approximation in the second term of the ML loss, we propose the following cost function, called {\it Soft Metric Learning} (SoftML) loss:
%\begin{equation*}
%\max \Big(0, l-d(\cdot)\Big) \approx \log\Big(1+\exp\big(l-d(\cdot)\big)\Big).
%\end{equation*}
%The same approximation turns the rectifier to the soft-plus function~\citep{dugas2001incorporating, goodfellow2016deep}. Then the new cost function is
\begin{equation}
\label{eq:J2}
%\begin{split}
J_{\text{SoftML}} = \sum_{i=1}^{N}\bigg(d_{y_i}(x^{(i)}) + \sum_{j \notin y_i}\log{\Big(1+\exp\big(\lambda-d_j(x^{(i)})\big)\Big)} \bigg) \\
% & = \sum_{i=1}^{N} -\log\Bigg(\frac{\exp\big(-d_{y_i}(x^{(i)}) \big)}{\prod_{j \notin y_i}\big(1+\exp(\lambda)\exp(-d_{j}(x^{(i)}))\big)} \Bigg).
%\end{split}
\end{equation}

This new cost, which does not differ much from the ML loss, help us find an output layer to map the outputs of the deep-RBF network to unnormalized probabilities. Let $O_i := \exp(-d_i(x))$~\footnote{ Note that $O_i$s are equivalent to the outputs of the deep-RBF network if the RBF units are $\phi(x) = \exp(-\norm{A^{T}x+b}_p^p)$.}, then one can observe that

\begin{equation}
J_{\text{SoftML}}  = -\log\prod_{i=1}^{N}\bigg( \frac{O_{y_i} (1+\exp(\lambda) O_{y_i})}{\prod_{j}(1+\exp(\lambda) O_j)} \bigg),
\label{eq:J2NLL}
\end{equation}
which is the negative log-likelihood when we use the following output layer after a deep-RBF network:

\begin{equation}
p(y=k | x) = \frac{O_k (1+\exp(\lambda) O_k)}{\prod_{j}(1+\exp(\lambda) O_j)}, \qquad  k \in \{1, 2, \dots, c\}.
\label{eq:OU}
\end{equation}
For the negative class, i.e., rejection, we use the following output which is equivalent to the smooth approximation of Eq.~\ref{eq:negLoss}:
\begin{equation}
p(y=c+1 | x) = \frac{1}{\prod_{j}(1+\exp(\lambda) O_j)}.
\label{eq:OUrej}
\end{equation}
%This output unit with negative log-likelihood for negative examples is equivalent to the loss $\sum_{j}\log{(1+\exp( \lambda-d_j(x^{(i)})))}$, which is the smooth approximation of Eq.~\ref{eq:negLoss}.

These outputs, which we call probabilities, are all positive and less than but not normalized to one. In the evaluation phase, we choose the output with the highest probability. It is easy to see that the output with the highest $p(y=k | x)$ is associated with the output with the smallest distance $d_k$. Furthermore, one can show that whenever $p(y=c+1 | x)$ is the biggest output, the smallest distance is greater than a threshold, i.e., $\min_{i \in \{1, \dots, c\}}d_i(x) >   T$. This threshold is
\begin{equation}
\label{eq:thr_lam}
T = -\log\bigg(\frac{-1+\sqrt{1+4\exp(\lambda_{\text{eval}})}}{2\exp(\lambda_{\text{eval}})}\bigg),
\end{equation}
which is always positive. Note that the parameter $\lambda_{\text{eval}}$ differ from $\lambda$ that we use in training. This parameter only affect the threshold, and we can regard it as another reparameterization of the threshold.
% for valid values of $\exp(\lambda)$, and is a strictly increasing function of $\exp(\lambda)$.

\subsubsection{Looking at LeNet5 loss from this probabilistic viewpoint}
We can treat the loss~\ref{eq:LeNet5V2}, used in the second strategy of LeNet5, as what we did above, and derive an output layer.
If we assume $O_i := \exp(-\hat{d}_i)$, then loss~\ref{eq:LeNet5V2} will be
\begin{equation}
J_{\text{LeNet5}}  = \frac{-1}{N}\log\prod_{i=1}^{N}\bigg( \frac{O_{y_i}}{\exp(-\lambda)+\sum_{j}O_j} \bigg).
\end{equation}
If we ignore the constant $1/N$, the above loss is the negative log-likelihood when we use the following output layer after a deep-RBF network with distance $\hat{d}_k(x) = \norm{f(x)-W_k}_2^2$:
\begin{equation}
\begin{split}
p(y=k | x) & = \frac{O_k}{\exp(-\lambda)+\sum_{j}O_j}, \qquad  k \in \{1, 2, \dots, c\}, \\
p(y=c+1 | x) & = \frac{\exp(-\lambda)}{\exp(-\lambda)+\sum_{j}O_j}.
\end{split}
\end{equation}
With this choice of output for the negative class, the rejection occurs when $\min_{i \in \{1, \dots, c\}}\hat{d}_i(x) > \lambda$, so the threshold is $\lambda$. The $c+1$ outputs are normalized to one. We can generalize the distance $\hat{d}_k(x)$ to Eq.~\ref{eq:distance} so that the above output layer can be employed for the general formulation of the deep-RBF networks. 
%In Section~\ref{sec:Result}, we compare these output units and observe that our proposed output unit is superior to the one derived from LeNet5.

%%%%%%%%%%%%%%%%%%%%%%%%%
\section{Experimental results}
\label{sec:Result}
In this section, by running some experiments on the MNIST dataset, we evaluate the deep-RBF networks and our approach for training them. First, we compare the performances of the ordinary CNN and the deep-RBF network trained with ML, SoftML, and LeNet5 cost functions. We then examine the robustness of the deep-RBF networks against adversarial attacks and compare it with the robustness of other networks. In all the experiments, we used the same 2-layer ConvNet as the feature extractor, whose activation function is ReLU. The dimensionality of the feature-space is ten, and two batch normalization layers were employed. We used the Adam optimization algorithm and initialized the wights of the network randomly with samples drawn from a standard normal distribution. While fixing the structure of the network, we chose some hyper-parameters including the learning rate, the parameter $\lambda$, and the number of epochs to achieve the best accuracy on the validation dataset. 

Firstly, we compare the classification accuracy of the deep-RBF network in different settings. As mentioned earlier, our method uses a generic definition of distance, appearing in~\ref{eq:distance}. For the norm of the distance, we consider two cases: $\ell_1$-norm and $\ell_2$-norm. Also, four different types of weight matrix $A_k$ are used. These types are the identity matrix ($A_k = I_{10\times 10}$), $A_k \in \mathbb{R}^{10\times 1}$, $A_k \in \mathbb{R}^{10\times 2}$, and $A_k \in \mathbb{R}^{10\times 5}$. The difference between the last three cases is the dimensionality of the output, which is equal to the rank of the covariance matrix $\Sigma_k=A_k A_k^T$. 
%Note that although the original distance of LeNet5 was the Euclidean distance ($\ell_2$-norm with the Identity weight matrix), we have substituted it with this general distance.
The classification accuracy of the ordinary CNN (i.e., the same 2-layer ConvNet with the softmax output layer) on the MNIST's test dataset is equal to $99.06$ percent. Table.~\ref{table:ACCs} displays this result for different settings of the deep-RBF network trained with three different cost functions: ML loss, SoftML loss, and LeNet5 loss. In all the experiments, the parameter $\lambda$ is equal to 550, except for the experiments with the identity weight matrix, wherein this parameter is 50. Note that we avoid rejecting in the evaluation phase because there are no negative examples both in the training and the evaluation. 

\begin{table}
\scriptsize
\caption{\small The classification accuracy of the deep-RBF network, with different distance forms, on MNIST. Three loss functions are employed for training. One can notice that the distance with the $\ell_1$-norm performs better, and our proposed cost functions typically yield to better classification accuracies than the cost function of LeNet5.}
\label{table:ACCs}
\begin{center}
\begin{small}
\begin{sc}
\begin{tabular}{lccccr}
\hline
%\abovespace\belowspace
 Deep-RBF distance form  & ML loss & SoftML loss & LeNet5 loss \\
\hline
%\abovespace
$\ell_1$-norm $\&$ Identity cov.   & 98.39$\%$ & 98.35$\%$ & 98.61$\%$ \\
$\ell_1$-norm $\&$ $1$ dim. output    & 99.21$\%$ & 99.21$\%$ & 98.17$\%$ \\
$\ell_1$-norm $\&$ $2$ dim.  output    & \bf{99.22}$\%$ & \bf{99.23}$\%$ & \bf{98.93}$\%$ \\
$\ell_1$-norm $\&$ $5$ dim.  output     & 99.18$\%$ & 99.20$\%$ & 98.80$\%$ \\
$\ell_2$-norm $\&$ Identity cov.   & 98.46$\%$ & 98.35$\%$ & 97.18$\%$ \\
$\ell_2$-norm $\&$ $1$ dim. output    & 98.49$\%$ & 99.03$\%$ & 88.13$\%$ \\
$\ell_2$-norm $\&$ $2$ dim. output     & 98.99$\%$ & 99.00$\%$ & 93.64$\%$ \\
$\ell_2$-norm $\&$ $5$ dim. output      & 98.88$\%$ & 98.96$\%$ & 95.58$\%$ \\
\hline
\end{tabular}
\end{sc}
\end{small}
\end{center}
\end{table}

The results in Table.~\ref{table:ACCs} indicate that the metric learning and soft metric learning losses do not differ a lot in the sense of classification accuracy; their negligible difference is due to the random initialization and our choice of some hyper-parameters.
They can also outperform the ordinary CNN, whose accuracy equals to $99.06\%$. Additionally, using the $\ell_1$-norm often yields to a better classification accuracy than the $\ell_2$-norm. It has been observed that the optimization algorithm converges faster when distances have the $\ell_1$-norm. In addition to the advantages provided by using a more general distance, our proposed methods have at least one other benefit over the ordinary LeNet5: that our proposed cost functions work better than the loss of LeNet5. One can acknowledge this fact by comparing the results in Table.~\ref{table:ACCs}. On the other hand, we observed in practice that the optimization procedure for our cost functions is faster and easier than that of LeNet5. The original method of LeNet5~\citep{lecun1998gradient} had tanh saturating units in its last layer, which we threw away because these units can deteriorate the optimization by causing the vanishing gradient problem.

As we mentioned in Section~\ref{sec:defdeep-RBF}, the authors of LeNet5 exploited some tricks to escape from local minimums and to make the network operate better; also the feature extractor network on their paper was bigger and more powerful than ours. Note that although the best accuracy of the deep-RBF network is somewhat near the state-of-the-art accuracy in the classification of MNIST, the experiments of this section were conducted only for the purpose of comparing the properties of these methods. By applying some tricks and choosing more powerful networks, we can improve the results. For example, several papers have proposed to utilize data augmentation methods, like {\it affine distortion} and {\it elastic distortion} to further increase the accuracy of the classifier~\citep{lecun1998gradient, simard2003best}. They may improve the performance of our method too.

\begin{figure}
\centering
        \includegraphics[totalheight=8.5cm, clip, trim=0cm 1.6cm 0cm 0.5cm]{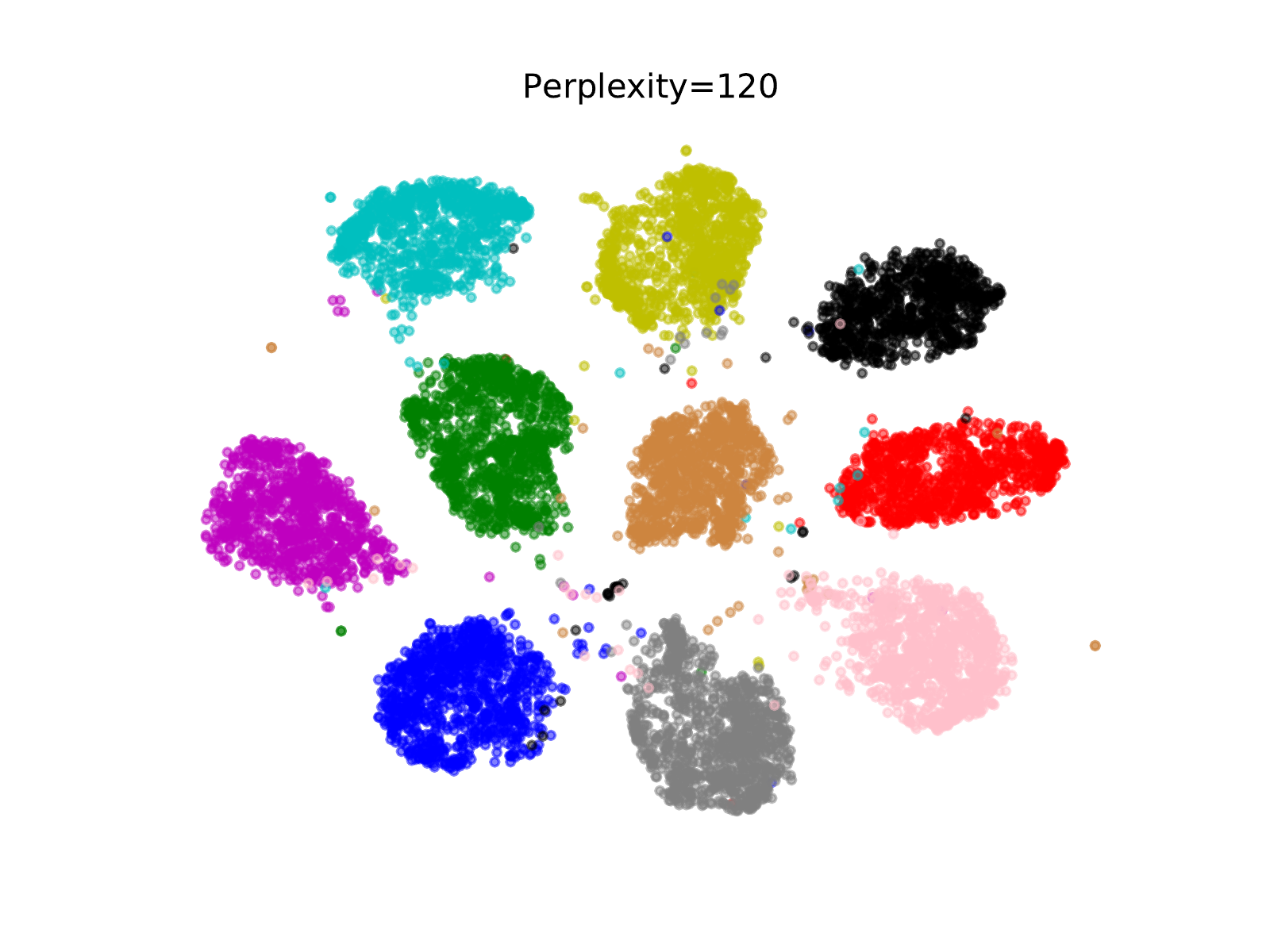}
    \caption{{\small Visualizing the feature space of the deep-RBF network using the t-SNE algorithm when the MNIST test dataset is passed through the network. Each color corresponds to a digit.}}
    \label{fig:tSNE}
\end{figure}

In order to provide some insight into the feature space of the deep-RBF network, i.e., the output units of the last layer before the classifier, we passed the test samples of MNIST through a deep-RBF network, trained with the ML loss, and recorded their feature vectors. Fig.~\ref{fig:tSNE} displays the results of the t-SNE dimensionality reduction algorithm (with perplexity $=120$) performing on them. As this figure illustrates, feature points near each other with high probability belong to the same category, which is a desirable property.

\begin{figure}
\centering
        \includegraphics[totalheight=9.6cm]{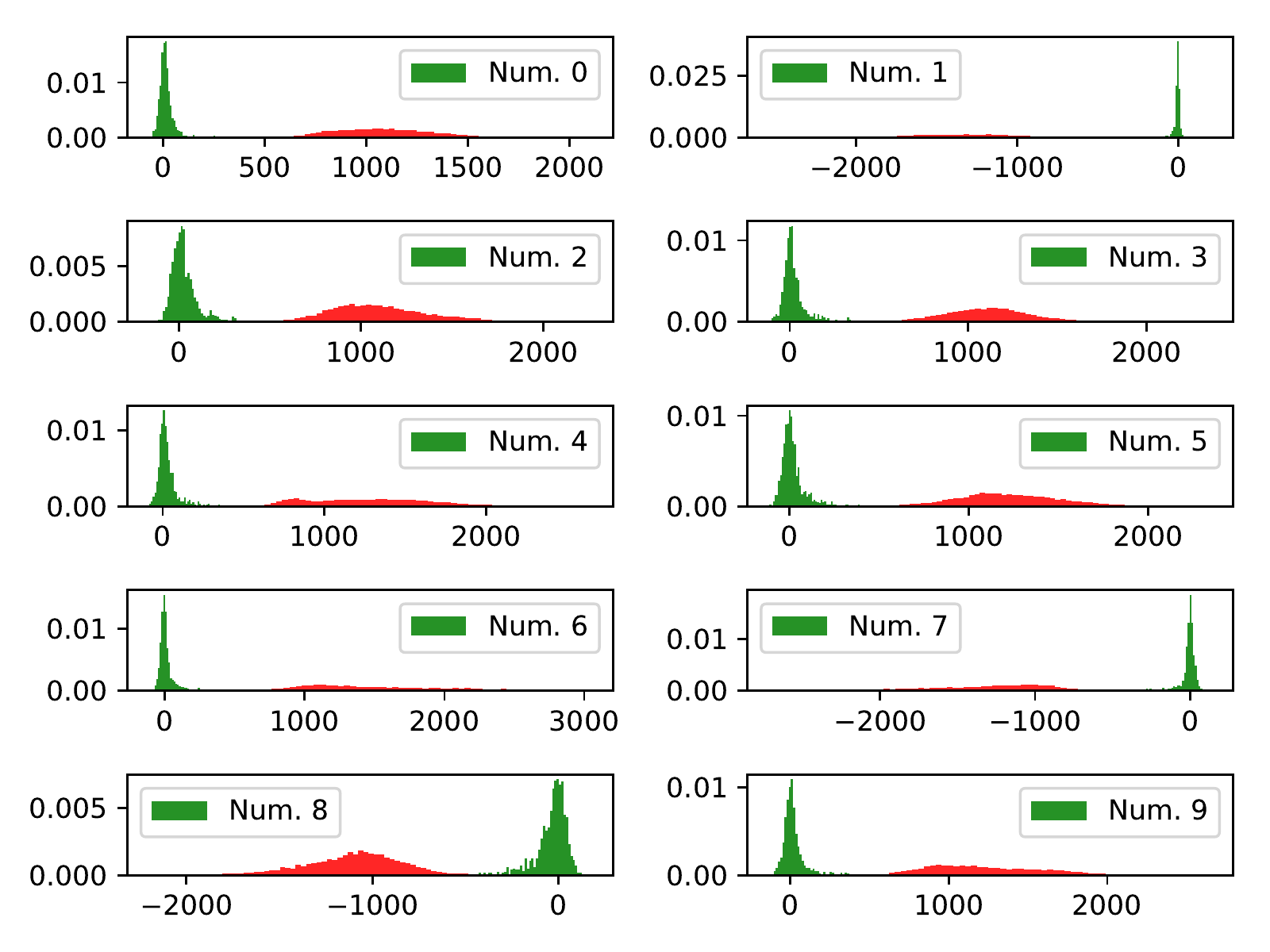}
    \caption{{\small The histogram of each category’s output when the MNIST test dataset is given to a trained deep-RBF network. In this network, the output of each category is one-dimensional, and its bias parameter is set to zero. The green color corresponds to the samples with the same label as each category whereas the red color corresponds to the samples with other labels. As shown in these plots, the green histograms center around the mean of the outputs, which is zero, and the red histograms lie only one side of them: the side to which the green histograms are slightly skewed. Moreover, the green histograms have lower variances than the red ones.}}
    \label{fig:1dimhist}
\end{figure}

Fig.~\ref{fig:1dimhist} is another visualization for the deep-RBF networks. It displays the histogram of the output of each category (i.e., $z_k = A_k^Tf(x)+b_k$) when the MNIST test dataset is given to a deep-RBF network which is trained with the ML cost function. Each category of the applied network has a one-dimensional output, and the distance has the $\ell_1$-norm. We set the bias term $b_k$ to zero (so did the mean) and observed that it does not affect the classification accuracy. There are two histograms in each plot that is associated with one digit. The green histogram corresponds to samples whose labels are the plot's category whereas the red one corresponds to the other samples. We can observe that the green histograms, which look like either the left-Gumbel or the right-Gumbel distributions, are around zero with lower variances than the red histograms. They are also slightly skewed to the side that the red histograms lie, however, we can draw a clear distinction between them.

%\begin{figure}[H]
%\centering
%        \includegraphics[totalheight=9cm]{figures/fig_nums.pdf}
%    \caption{{\small The.}}
%    \label{fig:1dimvis}
%\end{figure}

%\begin{figure}[H]
%\centering
%        \includegraphics[totalheight=10cm]{figures/fig_dist.pdf}
%    \caption{{\small The.}}
%    \label{fig:}
%\end{figure}

\begin{figure}
\centering
  \includegraphics[totalheight=8cm, clip, trim=0cm 0cm 0cm 1cm]{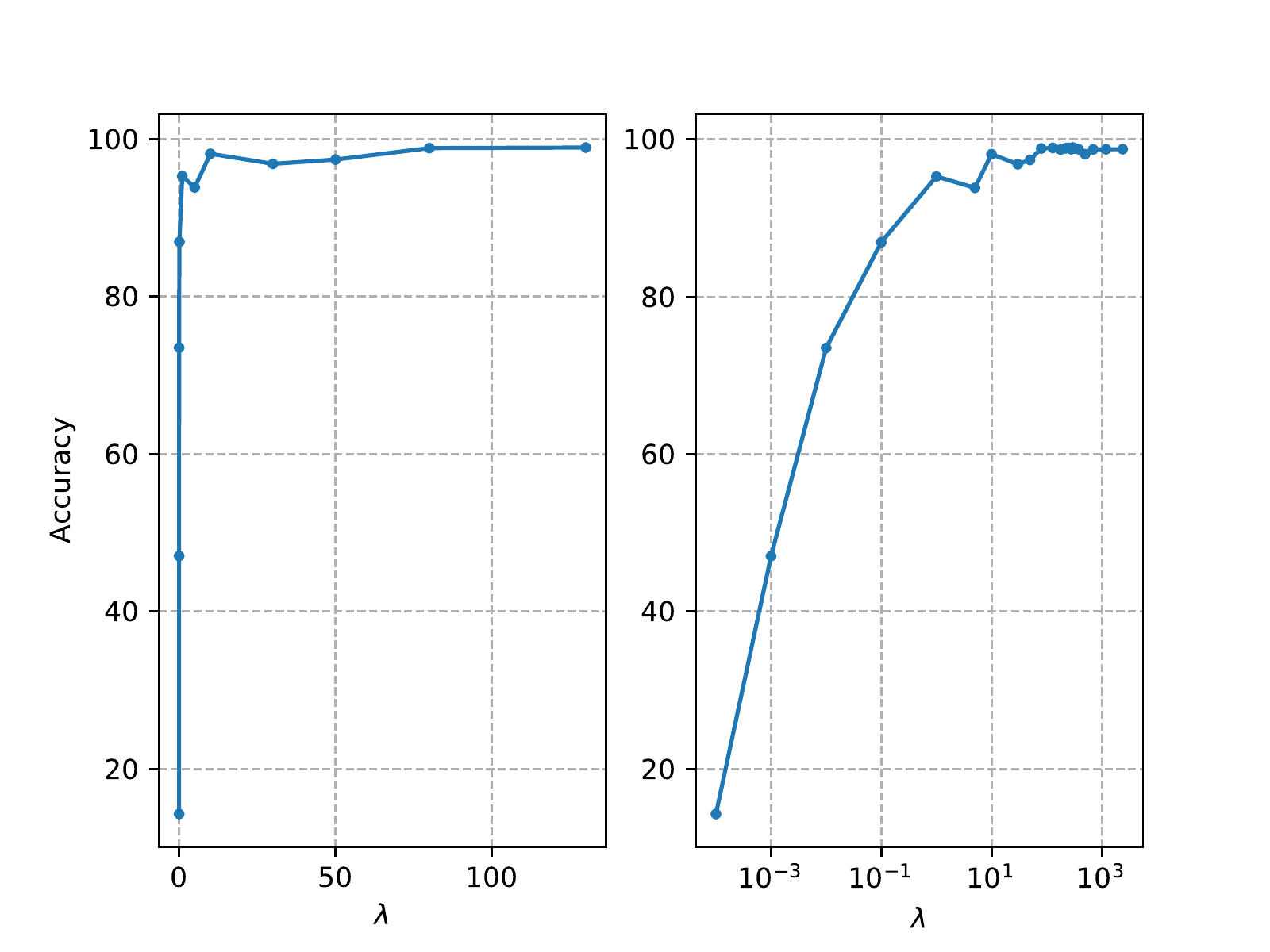}
\caption{\small The effect of the parameter $\lambda$ on the classification accuracy of the deep-RBF network when we avoid rejecting in the evaluation phases.}
\label{fig:lambda_train}
\end{figure}

As explained in Section~\ref{sec:learning}, the parameter  $\lambda$ only scales the optimal values of weights and biases in the last layer. It should not affect the classification accuracy if the initialization and learning method is also scaled with the same scaling factor. In reality, however, we have a fixed initialization and optimization method, and therefore, the parameter can affect the classification accuracy.
The plots in Fig.~\ref{fig:lambda_train} show the effect of the parameter $\lambda$ on the classification accuracy of the deep-RBF network when we avoid rejecting. In this plots, we use various values of $\lambda$ in training and then report the accuracy on the MNIST test dataset. In this figure, while the horizontal axis of the left plot is on a linear scale, that of the right one is on a logarithmic scale. As we see in the figure, the accuracy is flat for a wide range of $\lambda$ that shows the ease of its selection in practice.
%We can see that for the values of $\lambda$ over 100, we get a high classification accuracy.
%We can see that they both saturate gradually, and increasing $\lambda$ after a point neither improve nor worsen the classification accuracy. We have observed that the same  happens in every setting of the deep-RBF network and the deep-NCC network, except when the weight matrix is identity. Having this observation, we can use a large $\lambda$ parameter and no adjustment seems necessary. On the contrary, adjusting $\lambda$ in MoLeNet5 is much harder as its classification accuracy fluctuates a lot while the parameter $\lambda$ increases.

\begin{figure}
\centering
        \includegraphics[totalheight=9cm, clip, trim=0cm 0cm 0cm 1cm]{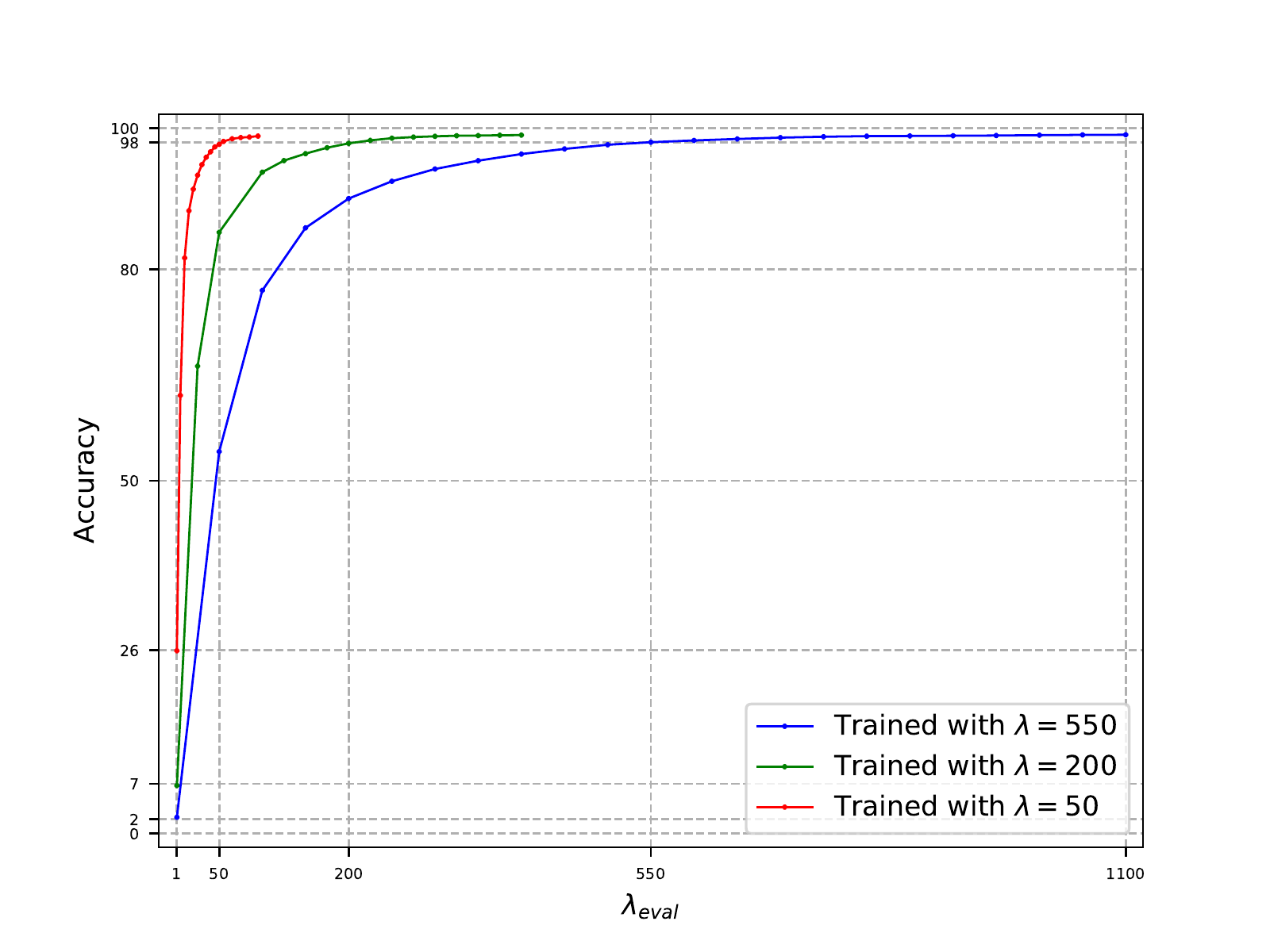}
    \caption{{\small The relationship between the parameter $\lambda_{\text{eval}}$, i.e., the threshold, and the classification accuracy of deep-RBF networks when the network rejects the samples with low confidence. We know from Eq.~\ref{eq:thr_lam} that $\lambda_{\text{eval}}$ is just another reparameterization of the threshold. Here, we used the SoftML cost function and trained the network with three different values of $\lambda$; each one is associated with one color in this figure. All the curves are increasing functions and converge to the accuracy of the method in which the rejection is avoided. An appropriate $\lambda_{\text{eval}}$ for having a particular accuracy is dependent on the value of $\lambda$ in the training phase. For example, these networks attain accuracies around $98\%$ by choosing the $\lambda_{\text{eval}}$ equal to the $\lambda$ with which the method is trained.}}
    \label{fig:lambda_eval}
\end{figure}

When our method is to reject in the evaluation phase, i.e., assigning the negative label to a sample with low confidence, we should determine the amount of threshold. If we use the SoftML cost with the probabilistic viewpoint, according to Eq.~\ref{eq:thr_lam}, the threshold only depends on the parameter $\lambda_{\text{eval}}$, which is another reparameterization of the threshold. Fig.~\ref{fig:lambda_eval} illustrates the relationship between $\lambda_{\text{eval}}$, i.e., the threshold, and the classification accuracy of the deep-RBF networks trained with different values of $\lambda$. We see in this figure that changing  $\lambda$ only scales the results. As expected, the curves are increasing functions of $\lambda_{\text{eval}}$. That happens because increasing the threshold causes the [hyper-]volume of the decision surfaces of the non-negative categories to go up; in other words, the rejection regions become smaller. Consequently, it yields to a growth in the classification accuracy. Moreover, each curve converges to a point where no rejection is performed as if we avoid rejecting.

%{\color{red} \#USE THRESHOLD for EVALUATION
%
%Now, we are going to discuss the effect of the parameter $\lambda$ on our proposed methods in both the training and evaluation phases. If we avoid rejecting in the evaluation phase, that is to say, no negative examples exist, adjusting $\lambda$ will be necessary only in training.
%
%
%
%
%As we mentioned earlier, if our method is to reject in the evaluation phase, i.e., assigning the negative label to a sample with low confidence, we should determine the amount of threshold. {\color{red} According to Eq.~\ref{eq:thr_lam}, this threshold is only dependent on the parameter $\lambda$; therefore, we should determine an appropriate value for $\lambda$ in the evaluation. Let us use the notation $\lambda_{\text{eval}}$ for this parameter, so we do not confuse it with the parameter $\lambda$ which is for training. Look at Fig.~\ref{fig:lambda_eval}; it illustrates the relationship between $\lambda_{\text{eval}}$ and the classification accuracy of deep-RBF networks when we consider rejecting. We have trained the network three times with different values of $\lambda$. As you can see, this parameter affects the performance of a given $\lambda_{\text{eval}}$ in the evaluation; the network trained with lower $\lambda$ changes more rapidly. We should consider this fact when we are going to select an appropriate $\lambda_{\text{eval}}$.}
%Another fact this figure reveals is that the curves are increasing functions of $\lambda_{\text{eval}}$.
%That happens because rising $\lambda_{\text{eval}}$ causes the [hyper-]volume of the decision surfaces of the non-negative categories to go up; in other words, the rejection regions become smaller. Consequently, it yields to a growth in the classification accuracy. Moreover, each curve converges to a point where no rejection is performed as if we avoid rejecting.

One can imply from Fig.~\ref{fig:lambda_eval} that as the threshold grows, the accuracy of our classifier in the absence of negative examples increases. However, as we see in the next subsection, increasing the threshold also makes the classifier more vulnerable to the adversarial attacks, and if the negative examples exist, the chance of assigning a non-negative category to a negative sample will become higher. Therefore, there is a trade-off, and we must find the right balance. In the tasks where negative examples exist, we can use the cross-validation to find this parameter. The criterion used in the cross-validation could be, for example, the classification accuracy when the negative class is regarded as an ordinary class. On the other hand, if the classifier is to cope with the adversarial attacks, we can choose the threshold after running adversarial attack methods on the trained network for various thresholds.

\subsection{Adversarial attacks}

In the introduction, we claimed that our method is resistant to adversarial attacks. Now, we want to support it by conducting some experiments. We run the attacks Fast Gradient Sign Method (FGSM), Iterative Gradient Sign Attack (IGSA),  Gradient Attack (GA), and L-BFGS on our the deep-RBF network, trained with ML, SoftML, and LeNet5 losses, and compare their results with those of ordinary CNN. The python package Foolbox [version 1.1.0]~\citep{rauber2017foolbox} with its default hyper-parameters has been used to perform the attacks.
%\footnote{ Note that its default cost function is the combination of softmax and NLL. To run the attacks on our methods, we must replace it with NLL. For Pytorch, in foolbox, go to models/pytorch.py and change ``ce = nn.CrossEntropyLoss()'' to ``ce = nn.NLLLoss()''.}
In all of the following experiments, we have used the first 100 samples of the MNIST test dataset as the input samples for the adversarial attacks, in order to achieve a fair comparison.

Before jumping into the results, let us talk about noise injection. In our experiments, we noticed that adding noise to the inputs during the deep-RBF network's training can improve the robustness of our methods substantially whereas it does not have the same effect on other methods like the ordinary CNN. We used additive isotropic Gaussian noise in the input: $\hat{x} = x+\epsilon, \epsilon \sim \mathcal{N}(0,\sigma^2I)$.
Two different perspectives can explain this phenomenon. From one point of view, towards which we incline, adding noise to inputs is a way to regularize the network, as it is in the denoising auto-encoders~\citep{vincent2010stacked}. In this point of view, we consider the amount of noise's variance as the degree of regularization. Note that weight decay differs from this regularization as it does not improve the resistance to the adversarial examples. In the other viewpoint, this is just a way for data augmentation, and it fills some empty regions of the space with virtual data.

Table~\ref{table:ACCs_Noisy} shows the classification accuracy of some networks trained with three different standard deviations: $0$, $0.2$, and $0.4$. As you can see, the noiseless cases have the best accuracies in our methods, but injecting noise with standard deviation $0.2$ improves the accuracy of the ordinary CNN slightly. We can further observe that the classification accuracies of the losses ML and SoftML are about the same, and they are both better than LeNet5. Note that, here, the structure of the deep-RBF network for different cost functions are the same: $\ell_1$-norm with two dimensional outputs.
%the only difference among MoLeNet5, deep-RBF network and deep-NCC is their objective function, so our proposed objective functions outperform the loss of MoLeNet5 w.r.t. the classification accuracy.

\begin{table}[h]
\scriptsize
\caption{\small The effect of noise injection on the classification accuracy on MNIST. The standard deviations of additive Gaussian noise are 0.2, and 0.4. The three first rows show the result of the same deep-RBF network with different cost functions.}
\label{table:ACCs_Noisy}
\begin{center}
\begin{small}
\begin{sc}
\begin{tabular}{cccccr}
\hline
%\abovespace\belowspace
  Methods & Noiseless & Noise $= 0.2$ & Noise $= 0.4$ \\
\hline
%\abovespace
ML loss ($\ell_1$-norm $\&$ $2$ Dim.)    & 99.22$\%$ & 99.02$\%$ & 98.74$\%$ \\
SoftML loss ($\ell_1$-norm $\&$ $2$ Dim.)  & 99.23$\%$ & 99.03$\%$ & 98.64$\%$ \\
LeNet5 loss ($\ell_1$-norm $\&$ $2$ Dim.)  & 98.93$\%$ & 98.79$\%$ & 98.38$\%$ \\
CNN    & 99.06$\%$ & 99.12$\%$ & 98.78$\%$ \\
\hline
\end{tabular}
\end{sc}
\end{small}
\end{center}
\end{table}

\subsubsection{Targeted FGSM, IGSA, and GA attacks}

Now we want to examine the robustness of the deep-RBF network to the targeted FGSM, IGSA, and GA attacks. In the experiments, deep-RBF networks and CNN have the same feature extractor. The distance form of the deep-RBF network has $\ell_1$-norm and a two-dimensional output; this setting achieves the best classification accuracy in Table~\ref{table:ACCs}. The threshold is chosen such that the classification accuracy of all the deep-RBF networks on the MNIST is either $98\%$ or $95\%$. The ordinary CNN does not have this parameter, so its accuracy is not changing. We have chosen different accuracies to show that lowering the threshold of rejection would increase the robustness of our methods. In order to analyze the effect of noise injection on the resistance of the methods, we considered two cases: the noiseless case and the case with noise, whose standard deviation is $0.2$. Furthermore, we employed $L^2$-regularization in the deep-RBF to compare the effect of which on robustness with noise injection.

\begin{figure}[h]
\centering
\begin{minipage}{.32\textwidth}
  \centering
\scalebox{0.7}{\begin{tikzpicture}
\begin{axis}[scale=1,
%axis lines*=left,
x axis line style={draw opacity=0},
width=8cm,
height=3.4in,
bar width = .3cm,
enlarge x limits=0.49,
    tick align=inside,
    tick style={draw=none},
    xtick={6,8,10},
    xticklabels={%
        \small{$98$},
        \small{$95$},
        \small{CNN}},
     xticklabel style = { font=\scriptsize, text width=2.2cm, align=center },
    ybar,
    ymin = -18,
    xmajorticks=false,
    ymax = 120,
ytick={0,  20,  40,  60, 80, 100},
ylabel={ },
    legend image code/.code={%
                    \draw[#1, draw=none] (0cm,-0.1cm) rectangle (0.4cm,0.17cm);
                },  
                legend style={
                    %draw=none, % ?
                    text depth=0pt,
                    at={(0.37,0.99)},
                    anchor=north west,
                    legend columns=2,
                    default spacing:
                    %column sep=1cm,
                    % The text "Legend:"
                    /tikz/column 2/.style={column sep=0pt,font=\bfseries},
                    %
                    % the space between legend image and text:
                    /tikz/every odd column/.append style={column sep=0cm},
                },
    ]

\definecolor{c1}{RGB}{141,211,199}
\definecolor{c2}{RGB}{255,255,179}
\definecolor{c3}{RGB}{190,186,218}
\definecolor{c4}{RGB}{251,128,114}
%\definecolor{c5}{RGB}{85,87,179}
\definecolor{c6}{RGB}{253,180,98}

\addplot[
    fill=c1,
    draw=black,
    point meta=y,
    every node near coord/.style={inner ysep=5pt},
    error bars/.cd,
        y dir=both,
        y explicit
] 
table [y] {
x      y  
6   83
8   61
};

\addplot[
    fill=c1,
    postaction={
        pattern=my crosshatch dots
    },
    draw=black,
    point meta=y,
    every node near coord/.style={inner ysep=5pt},
    error bars/.cd,
        y dir=both,
        y explicit
] 
table [y] {
x      y  
6   57
8   17
};

\addplot[
    fill=c2,
    draw=black,
    point meta=y,
    every node near coord/.style={inner ysep=5pt},
    error bars/.cd,
        y dir=both,
        y explicit
] 
table [y] {
x      y     
6   68
8   34   
};

\addplot[
    fill=c2,
    postaction={
        pattern=my crosshatch dots
    },
    draw=black,
    point meta=y,
    every node near coord/.style={inner ysep=5pt},
    error bars/.cd,
        y dir=both,
        y explicit
] 
table [y] {
x      y     
6   46   
8   15   
};

\addplot[
    fill=c3,
    draw=black,
    point meta=y,
    every node near coord/.style={inner ysep=5pt},
    error bars/.cd,
        y dir=both,
        y explicit
] 
table [y] {
x       y   
6   92 
8   72 
};

\addplot[
    fill=c3,
    postaction={
        pattern=my crosshatch dots
    },
    draw=black,
    point meta=y,
    every node near coord/.style={inner ysep=5pt},
    error bars/.cd,
        y dir=both,
        y explicit
] 
table [y] {
x       y   
6   82 
8   56 
};

\addplot[
    fill=c4,
    draw=black,
] 
table [y] {
x       y  
8.4   99
};

\addplot[
    fill=c4,
    postaction={
        pattern=my crosshatch dots
    },
    draw=black,
] 
table [y] {
x       y  
8.4   97
};

\draw [decorate,decoration={brace,amplitude=10pt,raise=4pt},yshift=0pt]
(6.57,-0.12) -- (4.88,-0.12) node [black,midway,yshift=-0.7cm] {
$98\%$};

\draw [decorate,decoration={brace,amplitude=10pt,raise=4pt},yshift=0pt]
(8.57,-0.12) -- (6.88,-0.12) node [black,midway,yshift=-0.7cm] {
$95\%$};

\draw[line width=0.3 mm] (0,0) -- (12,0);
\draw[line width=0.3 mm] (0,100) -- (12,100);
\draw[line width=0.3 mm] (4.835,100) -- (4.835,0);
\draw[line width=0.3 mm] (9.565,100) -- (9.565,0);

%\draw ({rel axis cs:0,0}|-{axis cs:0,0}) -- ({rel axis cs:1,0}|-{axis cs:0,0});
\end{axis}
\end{tikzpicture} }
  \subcaption{{\small FGSM}}
  \label{fig:dsfas}
\end{minipage}
\begin{minipage}{.32\textwidth}
  \centering
\scalebox{0.7}{\begin{tikzpicture}
\begin{axis}[scale=1,
%axis lines*=left,
x axis line style={draw opacity=0},
width=8cm,
height=3.4in,
bar width = .3cm,
enlarge x limits=0.49,
    tick align=inside,
    tick style={draw=none},
    xtick={6,8,10},
    xticklabels={%
        \small{$98$},
        \small{$95$},
        \small{CNN}},
     xticklabel style = { font=\scriptsize, text width=2.2cm, align=center },
    ybar,
    ymin = -18,
    xmajorticks=false,
    ymax = 120,
ytick={0,  20,  40,  60, 80, 100},
ylabel={ },
    legend image code/.code={%
                    \draw[#1, draw=none] (0cm,-0.1cm) rectangle (0.4cm,0.17cm);
                },  
                legend style={
                    %draw=none, % ?
                    text depth=0pt,
                    at={(0.37,0.99)},
                    anchor=north west,
                    legend columns=2,
                    default spacing:
                    %column sep=1cm,
                    % The text "Legend:"
                    /tikz/column 2/.style={column sep=0pt,font=\bfseries},
                    %
                    % the space between legend image and text:
                    /tikz/every odd column/.append style={column sep=0cm},
                },
    ]

\definecolor{c1}{RGB}{141,211,199}
\definecolor{c2}{RGB}{255,255,179}
\definecolor{c3}{RGB}{190,186,218}
\definecolor{c4}{RGB}{251,128,114}
%\definecolor{c5}{RGB}{85,87,179}
\definecolor{c6}{RGB}{253,180,98}

\addplot[
    fill=c1,
    draw=black,
    point meta=y,
    every node near coord/.style={inner ysep=5pt},
    error bars/.cd,
        y dir=both,
        y explicit
] 
table [y] {
x      y  
6   95
8   69
};

\addplot[
    fill=c1,
    postaction={
        pattern=my crosshatch dots
    },
    draw=black,
    point meta=y,
    every node near coord/.style={inner ysep=5pt},
    error bars/.cd,
        y dir=both,
        y explicit
] 
table [y] {
x      y  
6   51
8   9
};

\addplot[
    fill=c2,
    draw=black,
    point meta=y,
    every node near coord/.style={inner ysep=5pt},
    error bars/.cd,
        y dir=both,
        y explicit
] 
table [y] {
x      y     
6   93
8   57
};

\addplot[
    fill=c2,
    postaction={
        pattern=my crosshatch dots
    },
    draw=black,
    point meta=y,
    every node near coord/.style={inner ysep=5pt},
    error bars/.cd,
        y dir=both,
        y explicit
] 
table [y] {
x      y     
6   46
8   3   
};

\addplot[
    fill=c3,
    draw=black,
    point meta=y,
    every node near coord/.style={inner ysep=5pt},
    error bars/.cd,
        y dir=both,
        y explicit
] 
table [y] {
x       y   
6   80
8   58 
};

\addplot[
    fill=c3,
    postaction={
        pattern=my crosshatch dots
    },
    draw=black,
    point meta=y,
    every node near coord/.style={inner ysep=5pt},
    error bars/.cd,
        y dir=both,
        y explicit
] 
table [y] {
x       y   
6   62
8   34 
};

\addplot[
    fill=c4,
    draw=black,
] 
table [y] {
x       y  
8.4   100
}; 

\addplot[
    fill=c4,
    postaction={
        pattern=my crosshatch dots
    },
    draw=black,
] 
table [y] {
x       y  
8.4   100
}; 

\draw [decorate,decoration={brace,amplitude=10pt,raise=4pt},yshift=0pt]
(6.57,-0.12) -- (4.88,-0.12) node [black,midway,yshift=-0.7cm] {
$98\%$};

\draw [decorate,decoration={brace,amplitude=10pt,raise=4pt},yshift=0pt]
(8.57,-0.12) -- (6.88,-0.12) node [black,midway,yshift=-0.7cm] {
$95\%$};

\draw[line width=0.3 mm] (0,0) -- (12,0);
\draw[line width=0.3 mm] (0,100) -- (12,100);
\draw[line width=0.3 mm] (4.835,100) -- (4.835,0);
\draw[line width=0.3 mm] (9.565,100) -- (9.565,0);

%\draw ({rel axis cs:0,0}|-{axis cs:0,0}) -- ({rel axis cs:1,0}|-{axis cs:0,0});
\end{axis}
\end{tikzpicture} }
  \subcaption{{\small IGSA}}
  \label{fig:dsfas}
\end{minipage}%
\begin{minipage}{.32\textwidth}
  \centering
\scalebox{0.7}{\begin{tikzpicture}
\begin{axis}[scale=1,
%axis lines*=left,
x axis line style={draw opacity=0},
width=8cm,
height=3.4in,
bar width = .3cm,
enlarge x limits=0.49,
    tick align=inside,
    tick style={draw=none},
    xtick={6,8,10},
    xticklabels={%
        \small{$98$},
        \small{$95$},
        \small{CNN}},
     xticklabel style = { font=\scriptsize, text width=2.2cm, align=center },
    ybar,
    ymin = -18,
    xmajorticks=false,
    ymax = 120,
ytick={0,  20,  40,  60, 80, 100},
ylabel={ },
    legend image code/.code={%
                    \draw[#1, draw] (0cm,-0.1cm) rectangle (0.4cm,0.17cm);
                },  
                legend style={
                    %draw=none, % ?
                    text depth=0pt,
                    at={(0.02,1.02)},
                    anchor=north west,
                    legend columns=2,
                    default spacing:
                    %column sep=1cm,
                    % The text "Legend:"
                    /tikz/column 2/.style={column sep=0pt,font=\bfseries},
                    %
                    % the space between legend image and text:
                    /tikz/every odd column/.append style={column sep=0cm},
                },
    ]

\definecolor{c1}{RGB}{141,211,199}
\definecolor{c2}{RGB}{255,255,179}
\definecolor{c3}{RGB}{190,186,218}
\definecolor{c4}{RGB}{251,128,114}
%\definecolor{c5}{RGB}{85,87,179}
\definecolor{c6}{RGB}{253,180,98}

\addplot[
    fill=c1,
    draw=black,
    point meta=y,
    every node near coord/.style={inner ysep=5pt},
    error bars/.cd,
        y dir=both,
        y explicit
] 
table [y] {
x      y  
6   63
8   43
}; \addlegendentry{ SoftML-noiseless}

\addplot[
    fill=c1,
    postaction={
        pattern=my crosshatch dots
    },
    draw=black,
    point meta=y,
    every node near coord/.style={inner ysep=5pt},
    error bars/.cd,
        y dir=both,
        y explicit
] 
table [y] {
x      y  
6   52
8   20
}; \addlegendentry{ SoftML-with noise}

\addplot[
    fill=c2,
    draw=black,
    point meta=y,
    every node near coord/.style={inner ysep=5pt},
    error bars/.cd,
        y dir=both,
        y explicit
] 
table [y] {
x      y     
6   49
8   23
}; \addlegendentry{ ML-noiseless}

\addplot[
    fill=c2,
    postaction={
        pattern=my crosshatch dots
    },
    draw=black,
    point meta=y,
    every node near coord/.style={inner ysep=5pt},
    error bars/.cd,
        y dir=both,
        y explicit
] 
table [y] {
x      y     
6   49
8   17
}; \addlegendentry{ ML-with noise}

\addplot[
    fill=c3,
    draw=black,
    point meta=y,
    every node near coord/.style={inner ysep=5pt},
    error bars/.cd,
        y dir=both,
        y explicit
] 
table [y] {
x       y   
6   78
8   62 
}; \addlegendentry{ LeNet5-noiseless}

\addplot[
    fill=c3,
    postaction={
        pattern=my crosshatch dots
    },
    draw=black,
    point meta=y,
    every node near coord/.style={inner ysep=5pt},
    error bars/.cd,
        y dir=both,
        y explicit
] 
table [y] {
x       y   
6   72
8   32 
}; \addlegendentry{ LeNet5-with noise}

\addplot[
    fill=c4,
    draw=black,
] 
table [y] {
x       y  
8.4   47
}; \addlegendentry{ CNN-noiseless}

\addplot[
    fill=c4,
    postaction={
        pattern=my crosshatch dots
    },
    draw=black,
] 
table [y] {
x       y  
8.4   76
};  \addlegendentry{ CNN-with noise}

\draw [decorate,decoration={brace,amplitude=10pt,raise=4pt},yshift=0pt]
(6.57,-0.12) -- (4.88,-0.12) node [black,midway,yshift=-0.7cm] {
$98\%$};

\draw [decorate,decoration={brace,amplitude=10pt,raise=4pt},yshift=0pt]
(8.57,-0.12) -- (6.88,-0.12) node [black,midway,yshift=-0.7cm] {
$95\%$};

\draw[line width=0.3 mm] (0,0) -- (12,0);
\draw[line width=0.3 mm] (0,100) -- (12,100);
\draw[line width=0.3 mm] (4.835,100) -- (4.835,0);
\draw[line width=0.3 mm] (9.565,100) -- (9.565,0);

%\draw ({rel axis cs:0,0}|-{axis cs:0,0}) -- ({rel axis cs:1,0}|-{axis cs:0,0});
\end{axis}
\end{tikzpicture} }
  \subcaption{{\small GA}}
  \label{fig:jkhdas}
\end{minipage}
\caption{{\small The success rate of the targeted adversarial attacks FGSM, IGSA, and GA on the deep-RBF networks and CNN. In the experiments, the same deep-RBF network is trained with different cost functions: ML loss, SoftML loss, and LeNet5 loss. While in the noiseless cases, the training is ordinary, in the cases with noise, we inject an additive Gaussian noise into the training samples. As we can see, the deep-RBF network is more resistant to these attacks than CNN with softmax output layer, primarily when we train the deep-RBF network using the ML loss and noise injection.}}
\label{fig:sucessRateTargeted}
\end{figure}
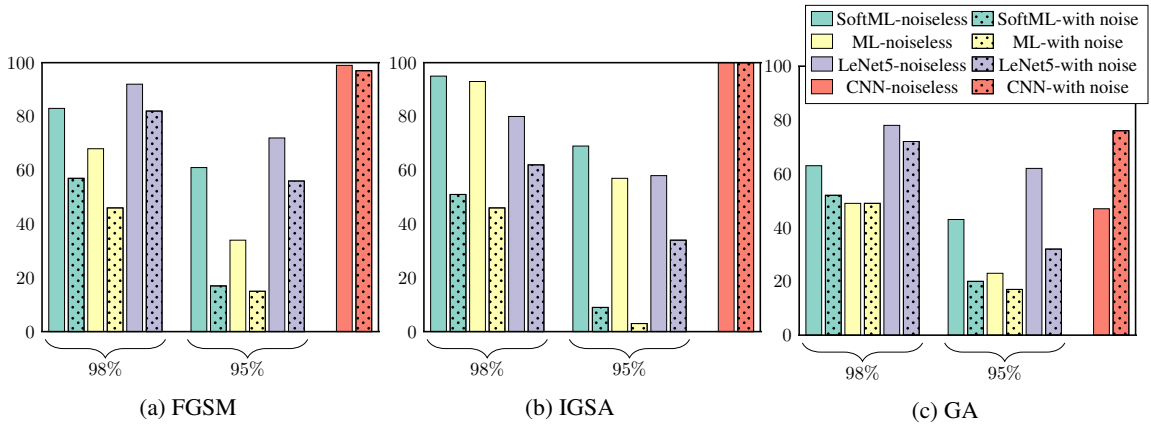

Fig.~\ref{fig:sucessRateTargeted} and Table~\ref{table:ADVs} show the results of the targeted adversarial attacks FGSM, IGSA, and GA. In these experiments, for every input image, we chose all the nine wrong classes, one by one, as the target and performed the attacks to fool the classifiers. Fig.~\ref{fig:sucessRateTargeted} displays the success rate of the attacks, which is the percentage of the input images which have been fooled at least in one target. In Table~\ref{table:ADVs}, MSpS stands for Mean Success per Sample and is the average number of targets with which each input sample is fooled. The Ratio of Confidences (RoC) expresses the ratio of the average confidence of the network for the correct class of the original examples to the average confidence of the network for the target class of adversarial examples, which were successful in fooling the network. The confidence, as usual, is the network's outputs, so for the ordinary CNN, it is the output of the softmax for the chosen class. Lastly, RMSD is the average Root Mean Squared Distortion between the original image $x$ and the distorted one $x'$. It is measured by $\sqrt{\sum{(x'_i-x_i)^2}/n}$ , where $n=784$.

As can be seen in Table~\ref{table:ADVs} and Fig.~\ref{fig:sucessRateTargeted}, the deep-RBF networks are much more resistant to adversarial examples than the ordinary CNN, especially for the FGSM and IGSA attacks. Besides, we can see that the cost functions ML and SoftML result in more robust classifiers than the cost of LeNet5. As expected, one consequence of decreasing the threshold for rejection, i.e., moving from the accuracy of $98\%$ to $95\%$, is increasing the resistance of the classifiers to adversarial examples. Its other consequence is declining the ratio of confidences, as Table~\ref{table:ADVs} reports. Moreover, the data in this table and figure indicate that noise injection can improve the robustness of deep-RBF network substantially, and this improvement occurs in every attack. However, this is not happening in the ordinary CNN; it implies that noise injection affects CNN and our methods differently.

\begin{table}
\scriptsize
\caption{\small The results of several targeted adversarial attacks on different settings of the deep-RBF networks and the ordinary CNN. MSpS stands for Mean Success per Sample. The Ratio of Confidences (RoC) expresses the ratio of the average confidence of the original examples to the average confidence of adversarial examples. Indeed, the RoC shows how much the confidences of the clean and adversarial examples differ. RMSD is the average Root Mean Squared Distortion between the original and the distorted images. The noiseless cases refer to the standard training, and ``with noise'' indicates that additive Gaussian noise is injected into the network during the training phase. To evaluate the effect of the threshold on the robustness, we run the experiments for the deep-RBF networks with two different accuracies. In other words, we find the thresholds such that the classification accuracy on the MNIST test dataset is either $98\%$ or $95\%$. As reported in the table, deep-RBF networks usually have smaller MSpS, bigger ROC, and bigger RMSD, all of which are signs of robustness. The gain is most significant when we train the network with ML loss and noise injection.}
\begin{center}
\begin{footnotesize}
\begin{tabular}{c|c|c|c|||c|c|c|||c|c|c|}
\cline{2-10}
                                                      & \multicolumn{3}{c|||}{{\bf FGSM}} & \multicolumn{3}{c|||}{{\bf IGSA}} & \multicolumn{3}{c|}{{\bf GA}} \\ \hline 
\multicolumn{1}{|c||}{{\bf Methods}}   &  \hspace{-0.23cm}  {\bf MSpS}  \hspace{-0.23cm}  &    {\bf RoC}   &    \hspace{-0.25cm} {\bf RMSD}  \hspace{-0.25cm}    &  \hspace{-0.23cm}  {\bf MSpS} \hspace{-0.23cm}   &    {\bf RoC}    &   \hspace{-0.25cm}  {\bf RMSD}  \hspace{-0.25cm}   &   \hspace{-0.23cm} {\bf MSpS}  \hspace{-0.23cm}  &    {\bf RoC}    &  \hspace{-0.25cm}   {\bf RMSD} \hspace{-0.25cm}    \\ \hline \hline
\multicolumn{1}{|c||}{{\bf SoftML loss} -- noiseless -- 98\%\,\,\,}     &    0.97    &   9e+116     &   0.229     &   4.86   &    11091    &   0.169    &   0.73      &   8e+116     &   0.203     \\ \hline
\multicolumn{1}{|c||}{{\bf SoftML loss} -- with noise -- 98\%}       &   0.67     &   6e+128      &    0.441     &   0.96     &    4.7e+8    &    0.214   &     0.65    &    1e+128    &   0.228     \\ \hline \hline
\multicolumn{1}{|c||}{{\bf ML loss} -- noiseless -- 98\%\,\,\,}    &    0.79    &    7e+116     &   0.242    &    3.93    &   206.9     &    0.175    &    0.6     &   8e+115     &   0.207     \\ \hline 
\multicolumn{1}{|c||}{{\bf ML loss} -- with noise -- 98\%}        &    0.48    &    6e+129     &    0.416   &    0.76    &   4950    &    0.213   &    0.54     &    5e+127    &    0.225    \\ \hline \hline
\multicolumn{1}{|c||}{{\bf LeNet5 loss} -- noiseless -- 98\%\,\,\,}    &    1.80    &    2.12    &    0.254   &   2.15     &    1.61    &    0.127   &    1.41     &    2.08    &    0.158    \\ \hline
\multicolumn{1}{|c||}{{\bf LeNet5 loss} -- with noise -- 98\%}      &    1.63    &    2.37     &    0.312     &    1.1    &    1.87    &    0.149     &     1.12    &    2.28    &    0.161    \\ \hline \hline \hline
\multicolumn{1}{|c||}{{\bf SoftML loss} -- noiseless -- 95\%\,\,\,}	     &    0.65    &    4e+66     &   0.294      &     1.41   &    103.6    &   0.186    &   0.45      &   1e+66     &   0.273     \\ \hline
\multicolumn{1}{|c||}{{\bf SoftML loss} -- with noise -- 95\%}       &    0.19    &    2e+69     &    0.516    &    0.09    &  1.5e+9      &    0.278   &    0.22     &   1e+68     &     0.226   \\ \hline \hline
\multicolumn{1}{|c||}{{\bf ML loss} -- noiseless -- 95\%\,\,\,}	   &    0.36    &    8e+65     &   0.264     &   1.09     &    238.9    &    0.178  &    0.25     &    6e+65    &    0.206    \\ \hline 
\multicolumn{1}{|c||}{{\bf ML loss} -- with noise -- 95\%}        &    0.15    &    1e+71     &   0.496      &    0.03    &     1e+23   &    0.230   &    0.18     &    6e+68    &    0.193    \\ \hline \hline
\multicolumn{1}{|c||}{{\bf LeNet5 loss} -- noiseless -- 95\%\,\,\,}     &    1.25    &    2.12    &    0.260      &   1.4     &    1.66    &    0.121    &     1.05     &    2.08    &    0.169    \\ \hline
\multicolumn{1}{|c||}{{\bf LeNet5 loss} -- with noise -- 95\%}        &    0.8    &    2.30     &    0.306     &    0.64    &    1.80    &    0.159    &     0.51    &    2.28    &    0.146    \\ \hline \hline \hline
\multicolumn{1}{|c||}{{\bf CNN} -- noiseless\,\,\,}  &    1.8    &    2.00    &    0.305    &    1.03    &   1.62     &    0.050   &    0.66     &    1.97    &    0.125    \\ \hline
\multicolumn{1}{|c||}{{\bf CNN} -- with noise}     &     2.18    &    2.26    &    0.323   &    1.16    &    1.82    &    0.102     &   1.32    &    2.16     &   0.161        \\ \hline
\end{tabular}
\end{footnotesize}
\end{center}
\label{table:ADVs}
\end{table}

Table~\ref{table:ADVs} reports another desirable property of our proposed cost functions for training the deep-RBF networks: that the ratios of the average confidences for ML and SoftML losses are remarkably larger than when the LeNet5 cost function is used. This difference in the ratios of confidences, which is also significant between deep-RBF network and CNN, means the few adversarial examples that are successful in fooling deep-RBF networks have very low confidences. The other fact showed by this table is that the ML loss yields to more robust classifiers than the SoftML loss.

 In order to compare the effect of noise injection and weight-decay regularization on the robustness of the deep-RBF networks, we trained the same deep-RBF network appearing in Table~\ref{table:ADVs} with ML loss and $L^2$-regularization (without noise injection). The classification accuracy of the network on the test dataset is $98.91$, and the regularization parameter is $0.05$. We also set the threshold to have two accuracies: $98\%$ and $95\%$. The result of this experiment appears in Table~\ref{table:ADVs2REG}. SR in this table stands for Success Rate. Comparing the results in Tables~\ref{table:ADVs2REG} with Table~\ref{table:ADVs} and Fig.~\ref{fig:sucessRateTargeted}, we can confirm that the weight decay regularization worsens the resistance of deep-RBF network, even worse than the noiseless case. Hence, noise injection acts differently from the weight-decay regularization on the deep-RBF networks.

\begin{table}
\setlength{\tabcolsep}{3.8pt}
\scriptsize
\caption{\small The results of targeted adversarial attacks FGSM, IGSA, and GA on the deep-RBF network when we employed ML loss with $L^2$-regularization for training. SR stands for Success rate, and MSpS stands for Mean Success per Sample. The Ratio of Confidences (RoC) expresses the ratio of the average confidence of the original examples to the average confidence of adversarial examples. RMSD is the average Root Mean Squared Distortion between the original and the distorted images.}
\begin{center}
\begin{footnotesize}
\begin{tabular}{c|c|c|c|c|||c|c|c|c|||c|c|c|c|}
\cline{2-13}
                                                      & \multicolumn{4}{c|||}{{\bf FGSM}} & \multicolumn{4}{c|||}{{\bf IGSA}} & \multicolumn{4}{c|}{{\bf GA}} \\ \hline 
\multicolumn{1}{|c||}{{\bf Methods}} &    {\bf  SR}     &  \hspace{-0.23cm}  {\bf MSpS}  \hspace{-0.23cm}  &    {\bf RoC}   &    \hspace{-0.25cm} {\bf RMSD}  \hspace{-0.25cm}  &     {\bf  SR}     &  \hspace{-0.23cm}  {\bf MSpS} \hspace{-0.23cm}   &    {\bf RoC}    &   \hspace{-0.25cm}  {\bf RMSD}  \hspace{-0.25cm}   &   {\bf  SR}     &   \hspace{-0.23cm} {\bf MSpS}  \hspace{-0.23cm}  &    {\bf RoC}    &  \hspace{-0.25cm}   {\bf RMSD} \hspace{-0.25cm}    \\ \hline \hline
\multicolumn{1}{|c||}{\hspace{-0.2cm} {\bf ML loss \& L2-reg} -- 98\%}          &     77\%    &    0.96    &    1e+135     &    0.224    &    99\%    &    3.42    &   8e+90    &    0.139    &   65\%    &    0.81     &    1e+135    &    0.195    \\ \hline
\multicolumn{1}{|c||}{\hspace{-0.2cm} {\bf ML loss \& L2-reg} -- 95\%}          &     45\%    &    0.46    &    6e+80     &    0.344    &    70\%    &    0.97    &   26.3    &    0.141    &   28\%    &    0.30     &    6e+80    &    0.262    \\ \hline
\end{tabular}
\end{footnotesize}
\end{center}
\label{table:ADVs2REG}
\end{table}

Now, we want to evaluate the cross-model generalization of the adversarial examples between our methods and the ordinary CNN. In these experiments, for the sake of compactness, we only examine the adversarial examples transferable between CNN and the deep-RBF network that is trained with ML loss because the results of other costs are approximately the same. The performed attacks, like Table~\ref{table:ADVs}, are the targeted version of FGSM, IGSA, and GA. Tables~\ref{table:cross_FGSM},~\ref{table:cross_IGSA} and~\ref{table:cross_GA}  show the results of  FGSM, IGSA, and GA attacks, respectively. Each row of the tables corresponds to the adversarial examples of a model, and the columns indicate the error induced by distorted examples which are fed to one given model. While the entries report the percentage of the adversarial examples transferable between the two models, the number between the parentheses is the percentage of distorted examples deep-RBF rejects.

\begin{table}[h]
\setlength{\tabcolsep}{4.5pt}
\scriptsize
\caption{\small The cross-model generalization of adversarial examples when the targeted FGSM attack runs. Each row corresponds to the adversarial examples of one model, and every column indicates the error induced by distorted examples which are fed to one given model. The entries report the percentage of the adversarial examples transferable between the two models, and the number between the parentheses is the percentage of rejection. Here, the cost function of the deep-RBF networks is the ML loss; we trained them with and without noise injection. While the Deep-RBF network rejects the adversarial examples which received low confidence, CNN does not.}
\label{table:cross_FGSM}
\begin{center}
\begin{small}
\begin{tabular}{c|c|c|c|c|c|c|}
\cline{2-7}
                       & {\begin{tabular}[c]{@{}c@{}}{\bf CNN}\\ noiseless\end{tabular}} & {\begin{tabular}[c]{@{}c@{}}{\bf CNN}\\ with noise\end{tabular}} & {\begin{tabular}[c]{@{}c@{}}{\bf Deep-RBF}\\ 98\% - noiseless\end{tabular}} & {\begin{tabular}[c]{@{}c@{}}{\bf Deep-RBF}\\ 98\% - with noise\end{tabular}} & {\begin{tabular}[c]{@{}c@{}}{\bf Deep-RBF}\\ 95\% - noiseless\end{tabular}} &{\begin{tabular}[c]{@{}c@{}}{\bf Deep-RBF}\\ 95\% - with noise\end{tabular}} \\ \hline 
\multicolumn{1}{|c||}{\begin{tabular}[c]{@{}c@{}}{\bf CNN}\\ noiseless\end{tabular}} & \begin{tabular}[c]{@{}c@{}}100\%\\ —\end{tabular} & \begin{tabular}[c]{@{}c@{}}39.4\%\\ —\end{tabular} & \begin{tabular}[c]{@{}c@{}}23.8\%\\ (34.4\%)\end{tabular} & \begin{tabular}[c]{@{}c@{}}4.44\%\\ (50.0\%)\end{tabular} & \begin{tabular}[c]{@{}c@{}}10.5\%\\ (57.2\%)\end{tabular} & \begin{tabular}[c]{@{}c@{}}0.55\%\\ (58.8\%)\end{tabular} \\ \hline
\multicolumn{1}{|c||}{\begin{tabular}[c]{@{}c@{}}{\bf CNN}\\ with noise\end{tabular}} & \begin{tabular}[c]{@{}c@{}}54.1\%\\ —\end{tabular} & \begin{tabular}[c]{@{}c@{}}100\%\\ —\end{tabular} & \begin{tabular}[c]{@{}c@{}}49.5\%\\ (28.4\%)\end{tabular} & \begin{tabular}[c]{@{}c@{}}9.63\%\\ (61.0\%)\end{tabular} & \begin{tabular}[c]{@{}c@{}}23.8\%\\ (60.5\%)\end{tabular} & \begin{tabular}[c]{@{}c@{}}2.29\%\\ (79.3\%)\end{tabular} \\ \hline
\multicolumn{1}{|c||}{\begin{tabular}[c]{@{}c@{}}{\bf Deep-RBF}\\ 98\% - noiseless\end{tabular}} & \begin{tabular}[c]{@{}c@{}}27.8\%\\ —\end{tabular} & \begin{tabular}[c]{@{}c@{}}15.1\%\\ —\end{tabular} & \begin{tabular}[c]{@{}c@{}}100\%\\ (0\%)\end{tabular} & \begin{tabular}[c]{@{}c@{}}0.00\%\\ (56.9\%)\end{tabular} & \begin{tabular}[c]{@{}c@{}}13.9\%\\ (86.0\%)\end{tabular} & \begin{tabular}[c]{@{}c@{}}0.00\%\\ (77.2\%)\end{tabular} \\  \hline
\multicolumn{1}{|c||}{\begin{tabular}[c]{@{}c@{}}{\bf Deep-RBF}\\ 98\% - with noise\end{tabular}} & \begin{tabular}[c]{@{}c@{}}64.5\%\\ —\end{tabular} & \begin{tabular}[c]{@{}c@{}}66.6\%\\ —\end{tabular} & \begin{tabular}[c]{@{}c@{}}56.2\%\\ (37.5\%)\end{tabular} & \begin{tabular}[c]{@{}c@{}}100\%\\ (0\%)\end{tabular} & \begin{tabular}[c]{@{}c@{}}31.2\%\\ (64.5\%)\end{tabular} & \begin{tabular}[c]{@{}c@{}}0\%\\ (100\%)\end{tabular} \\ \hline
\multicolumn{1}{|c||}{\begin{tabular}[c]{@{}c@{}}{\bf Deep-RBF}\\ 95\% - noiseless\end{tabular}} & \begin{tabular}[c]{@{}c@{}}19.4\%\\ —\end{tabular} & \begin{tabular}[c]{@{}c@{}}19.4\%\\ —\end{tabular} & \begin{tabular}[c]{@{}c@{}}100\%\\ (0\%)\end{tabular} & \begin{tabular}[c]{@{}c@{}}0.00\%\\ (69.4\%)\end{tabular} & \begin{tabular}[c]{@{}c@{}}100\%\\ (0\%)\end{tabular} & \begin{tabular}[c]{@{}c@{}}0.00\%\\ (77.7\%)\end{tabular} \\ \hline
\multicolumn{1}{|c||}{\begin{tabular}[c]{@{}c@{}}{\bf Deep-RBF}\\ 95\% - with noise\end{tabular}} & \begin{tabular}[c]{@{}c@{}}86.6\%\\ —\end{tabular} & \begin{tabular}[c]{@{}c@{}}86.6\%\\ —\end{tabular} & \begin{tabular}[c]{@{}c@{}}66.6\%\\ (33.3\%)\end{tabular} & \begin{tabular}[c]{@{}c@{}}100\%\\ (0\%)\end{tabular} & \begin{tabular}[c]{@{}c@{}}40.0\%\\ (60.0\%)\end{tabular} & \begin{tabular}[c]{@{}c@{}}100\%\\ (0\%)\end{tabular} \\ \hline
\end{tabular}
\end{small}
\end{center}
\end{table}

\begin{table}[h]
\setlength{\tabcolsep}{4.5pt}
\scriptsize
\caption{\small The cross-model generalization of adversarial examples when the targeted IGSA attack runs.}
\label{table:cross_IGSA}
\begin{center}
\begin{small}
\begin{tabular}{c|c|c|c|c|c|c|}
\cline{2-7}
                       & {\begin{tabular}[c]{@{}c@{}}{\bf CNN}\\ noiseless\end{tabular}} & {\begin{tabular}[c]{@{}c@{}}{\bf CNN}\\ with noise\end{tabular}} & {\begin{tabular}[c]{@{}c@{}}{\bf Deep-RBF}\\ 98\% - noiseless\end{tabular}} & {\begin{tabular}[c]{@{}c@{}}{\bf Deep-RBF}\\ 98\% - with noise\end{tabular}} & {\begin{tabular}[c]{@{}c@{}}{\bf Deep-RBF}\\ 95\% - noiseless\end{tabular}} &{\begin{tabular}[c]{@{}c@{}}{\bf Deep-RBF}\\ 95\% - with noise\end{tabular}} \\ \hline 
\multicolumn{1}{|c||}{\begin{tabular}[c]{@{}c@{}}{\bf CNN}\\ noiseless\end{tabular}} & \begin{tabular}[c]{@{}c@{}}100\%\\ —\end{tabular} & \begin{tabular}[c]{@{}c@{}}1.94\%\\ —\end{tabular} & \begin{tabular}[c]{@{}c@{}}0.00\%\\ (3.88\%)\end{tabular} & \begin{tabular}[c]{@{}c@{}}0.00\%\\ (3.88\%)\end{tabular} & \begin{tabular}[c]{@{}c@{}}0.00\%\\ (5.82\%)\end{tabular} & \begin{tabular}[c]{@{}c@{}}0.00\%\\ (92.2\%)\end{tabular} \\ \hline
\multicolumn{1}{|c||}{\begin{tabular}[c]{@{}c@{}}{\bf CNN}\\ with noise\end{tabular}} & \begin{tabular}[c]{@{}c@{}}14.6\%\\ —\end{tabular} & \begin{tabular}[c]{@{}c@{}}100\%\\ —\end{tabular} & \begin{tabular}[c]{@{}c@{}}9.48\%\\ (8.62\%)\end{tabular} & \begin{tabular}[c]{@{}c@{}}0.00\%\\ (15.5\%)\end{tabular} & \begin{tabular}[c]{@{}c@{}}3.44\%\\ (34.4\%)\end{tabular} & \begin{tabular}[c]{@{}c@{}}0.00\%\\ (22.4\%)\end{tabular} \\ \hline
\multicolumn{1}{|c||}{\begin{tabular}[c]{@{}c@{}}{\bf Deep-RBF}\\ 98\% - noiseless\end{tabular}} & \begin{tabular}[c]{@{}c@{}}19.1\%\\ —\end{tabular} & \begin{tabular}[c]{@{}c@{}}11.9\%\\ —\end{tabular} & \begin{tabular}[c]{@{}c@{}}100\%\\ (0\%)\end{tabular} & \begin{tabular}[c]{@{}c@{}}2.80\%\\ (28.2\%)\end{tabular} & \begin{tabular}[c]{@{}c@{}}79.6\%\\ (20.3\%)\end{tabular} & \begin{tabular}[c]{@{}c@{}}0.51\%\\ (49.1\%)\end{tabular} \\  \hline
\multicolumn{1}{|c||}{\begin{tabular}[c]{@{}c@{}}{\bf Deep-RBF}\\ 98\% - with noise\end{tabular}} & \begin{tabular}[c]{@{}c@{}}43.4\%\\ —\end{tabular} & \begin{tabular}[c]{@{}c@{}}39.4\%\\ —\end{tabular} & \begin{tabular}[c]{@{}c@{}}39.4\%\\ (15.7\%)\end{tabular} & \begin{tabular}[c]{@{}c@{}}100\%\\ (0\%)\end{tabular} & \begin{tabular}[c]{@{}c@{}}22.3\%\\ (47.3\%)\end{tabular} & \begin{tabular}[c]{@{}c@{}}55.2\%\\ (44.7\%)\end{tabular} \\ \hline
\multicolumn{1}{|c||}{\begin{tabular}[c]{@{}c@{}}{\bf Deep-RBF}\\ 95\% - noiseless\end{tabular}} & \begin{tabular}[c]{@{}c@{}}13.7\%\\ —\end{tabular} & \begin{tabular}[c]{@{}c@{}}11.9\%\\ —\end{tabular} & \begin{tabular}[c]{@{}c@{}}100\%\\ (0\%)\end{tabular} & \begin{tabular}[c]{@{}c@{}}1.83\%\\ (25.6\%)\end{tabular} & \begin{tabular}[c]{@{}c@{}}100\%\\ (0\%)\end{tabular} & \begin{tabular}[c]{@{}c@{}}0.91\%\\ (44.9\%)\end{tabular} \\ \hline
\multicolumn{1}{|c||}{\begin{tabular}[c]{@{}c@{}}{\bf Deep-RBF}\\ 95\% - with noise\end{tabular}} & \begin{tabular}[c]{@{}c@{}}100\%\\ —\end{tabular} & \begin{tabular}[c]{@{}c@{}}66.6\%\\ —\end{tabular} & \begin{tabular}[c]{@{}c@{}}33.3\%\\ (66.6\%)\end{tabular} & \begin{tabular}[c]{@{}c@{}}100\%\\ (0\%)\end{tabular} & \begin{tabular}[c]{@{}c@{}}0\%\\ (100\%)\end{tabular} & \begin{tabular}[c]{@{}c@{}}100\%\\ (0\%)\end{tabular} \\ \hline
\end{tabular}
\end{small}
\end{center}
\end{table}

\begin{table}[H]
\setlength{\tabcolsep}{4.5pt}
\scriptsize
\caption{\small The cross-model generalization of adversarial examples when the targeted GA attack runs.}
\label{table:cross_GA}
\begin{center}
\begin{small}
\begin{tabular}{c|c|c|c|c|c|c|}
\cline{2-7}
                       & {\begin{tabular}[c]{@{}c@{}}{\bf CNN}\\ noiseless\end{tabular}} & {\begin{tabular}[c]{@{}c@{}}{\bf CNN}\\ with noise\end{tabular}} & {\begin{tabular}[c]{@{}c@{}}{\bf Deep-RBF}\\ 98\% - noiseless\end{tabular}} & {\begin{tabular}[c]{@{}c@{}}{\bf Deep-RBF}\\ 98\% - with noise\end{tabular}} & {\begin{tabular}[c]{@{}c@{}}{\bf Deep-RBF}\\ 95\% - noiseless\end{tabular}} &{\begin{tabular}[c]{@{}c@{}}{\bf Deep-RBF}\\ 95\% - with noise\end{tabular}} \\ \hline 
\multicolumn{1}{|c||}{\begin{tabular}[c]{@{}c@{}}{\bf CNN}\\ noiseless\end{tabular}} & \begin{tabular}[c]{@{}c@{}}100\%\\ —\end{tabular} & \begin{tabular}[c]{@{}c@{}}22.7\%\\ —\end{tabular} & \begin{tabular}[c]{@{}c@{}}16.6\%\\ (18.1\%)\end{tabular} & \begin{tabular}[c]{@{}c@{}}6.06\%\\ (24.2\%)\end{tabular} & \begin{tabular}[c]{@{}c@{}}3.03\%\\ (37.8\%)\end{tabular} & \begin{tabular}[c]{@{}c@{}}1.51\%\\ (39.4\%)\end{tabular} \\ \hline
\multicolumn{1}{|c||}{\begin{tabular}[c]{@{}c@{}}{\bf CNN}\\ with noise\end{tabular}} & \begin{tabular}[c]{@{}c@{}}49.2\%\\ —\end{tabular} & \begin{tabular}[c]{@{}c@{}}100\%\\ —\end{tabular} & \begin{tabular}[c]{@{}c@{}}31.8\%\\ (12.8\%)\end{tabular} & \begin{tabular}[c]{@{}c@{}}18.2\%\\ (29.5\%)\end{tabular} & \begin{tabular}[c]{@{}c@{}}12.1\%\\ (46.9\%)\end{tabular} & \begin{tabular}[c]{@{}c@{}}2.27\%\\ (58.3\%)\end{tabular} \\ \hline
\multicolumn{1}{|c||}{\begin{tabular}[c]{@{}c@{}}{\bf Deep-RBF}\\ 98\% - noiseless\end{tabular}} & \begin{tabular}[c]{@{}c@{}}36.6\%\\ —\end{tabular} & \begin{tabular}[c]{@{}c@{}}30.0\%\\ —\end{tabular} & \begin{tabular}[c]{@{}c@{}}100\%\\ (0\%)\end{tabular} & \begin{tabular}[c]{@{}c@{}}6.66\%\\ (35.0\%)\end{tabular} & \begin{tabular}[c]{@{}c@{}}16.6\%\\ (83.3\%)\end{tabular} & \begin{tabular}[c]{@{}c@{}}1.66\%\\ (68.3\%)\end{tabular} \\  \hline
\multicolumn{1}{|c||}{\begin{tabular}[c]{@{}c@{}}{\bf Deep-RBF}\\ 98\% - with noise\end{tabular}} & \begin{tabular}[c]{@{}c@{}}68.5\%\\ —\end{tabular} & \begin{tabular}[c]{@{}c@{}}61.1\%\\ —\end{tabular} & \begin{tabular}[c]{@{}c@{}}51.8\%\\ (27.7\%)\end{tabular} & \begin{tabular}[c]{@{}c@{}}100\%\\ (0\%)\end{tabular} & \begin{tabular}[c]{@{}c@{}}27.7\%\\ (59.2\%)\end{tabular} & \begin{tabular}[c]{@{}c@{}}3.70\%\\ (96.3\%)\end{tabular} \\ \hline
\multicolumn{1}{|c||}{\begin{tabular}[c]{@{}c@{}}{\bf Deep-RBF}\\ 95\% - noiseless\end{tabular}} & \begin{tabular}[c]{@{}c@{}}36.0\%\\ —\end{tabular} & \begin{tabular}[c]{@{}c@{}}32.0\%\\ —\end{tabular} & \begin{tabular}[c]{@{}c@{}}100\%\\ (0\%)\end{tabular} & \begin{tabular}[c]{@{}c@{}}8.00\%\\ (40.0\%)\end{tabular} & \begin{tabular}[c]{@{}c@{}}100\%\\ (0\%)\end{tabular} & \begin{tabular}[c]{@{}c@{}}8.00\%\\ (60.0\%)\end{tabular} \\ \hline
\multicolumn{1}{|c||}{\begin{tabular}[c]{@{}c@{}}{\bf Deep-RBF}\\ 95\% - with noise\end{tabular}} & \begin{tabular}[c]{@{}c@{}}100\%\\ —\end{tabular} & \begin{tabular}[c]{@{}c@{}}77.7\%\\ —\end{tabular} & \begin{tabular}[c]{@{}c@{}}77.7\%\\ (16.6\%)\end{tabular} & \begin{tabular}[c]{@{}c@{}}100\%\\ (0\%)\end{tabular} & \begin{tabular}[c]{@{}c@{}}72.2\%\\ (27.8\%)\end{tabular} & \begin{tabular}[c]{@{}c@{}}100\%\\ (0\%)\end{tabular} \\ \hline
\end{tabular}
\end{small}
\end{center}
\end{table}

It has been reported in lots of papers that different networks will misclassify a relatively large fraction of adversarial examples; in other words, adversarial examples usually generalize across different architectures~\citep{szegedy2013intriguing}. \citet{goodfellow2014explaining} have conjectured that this happens due to the highly linear nature of deep networks. On the other hand, our experiments regarding the cross-model generalization show that our proposed methods have a relatively low number of adversarial examples transferable between our methods and the ordinary CNN. We can see in tables~\ref{table:cross_FGSM},~\ref{table:cross_IGSA}, and~\ref{table:cross_GA} that the found adversarial examples of CNN do not often fool our models.
And as the threshold decreases, our methods are fooled less. Additionally, the found adversarial examples for the deep-RBF networks fool CNN more commonly, especially if the threshold is low and we use noise injection.

As we can see in these tables, the deep-RBF network has a low number of transferable adversarial examples between the noiseless case and the case with noise injection. We have observed that the cross-hyperparameter generalization of our methods is lower than the ordinary CNN, and even observed that two deep-RBF networks without noise and with precisely similar hyperparameters can have different adversarial examples just because of  the random initialization in their training. Therefore, the decision surfaces of the trained deep-RBF networks are also dependent on the initialization.

In all of these tables, the adversarial examples of the deep-RBF network with the accuracy of $95\%$ always fool the same network with $98\%$ accuracy because these are the same network with the same initialization and training; their only difference is the amount of threshold. Another observation one can make is that the adversarial examples found for the noiseless deep-RBF are often unable to fool its counterpart with noise injection; the reverse occurs less frequently. When interpreting these tables, note that the remaining percentage of the mutual adversarial examples and the rejected ones are the percentages of the adversarial examples that the other model does not even reject them because of its tiny distortions. We can observe that this amount is almost large for the adversarial examples of CNN when we feed them to the deep-RBF network with the high threshold, i.e., the one with $98\%$ accuracy, especially for the attacks IGSA and GA.

\subsubsection{Untargeted attacks}

\begin{figure}[h]
\centering
\begin{minipage}{.24\textwidth}
  \centering
\scalebox{0.72}{\begin{tikzpicture}
\begin{axis}[scale=1,
%axis lines*=left,
x axis line style={draw opacity=0},
width=6.1cm,
height=3.4in,
bar width = .3cm,
enlarge x limits=0.49,
    tick align=inside,
    tick style={draw=none},
    xtick={6,8,10},
    xticklabels={%
        \small{$98$},
        \small{$95$},
        \small{CNN}},
     xticklabel style = { font=\scriptsize, text width=2.2cm, align=center },
    ybar=1,
    ymin = -18,
    xmajorticks=false,
    ymax = 120,
ytick={0,  20,  40,  60, 80, 100},
ylabel={ },
    legend image code/.code={%
                    \draw[#1, draw] (0cm,-0.1cm) rectangle (0.4cm,0.17cm);
                },  
                legend style={
                    %draw=none, % ?
                    text depth=0pt,
                    at={(0.04,0.99)},
                    anchor=north west,
                    legend columns=6,
                    default spacing:
                    %column sep=1cm,
                    % The text "Legend:"
                    /tikz/column 2/.style={column sep=0pt,font=\bfseries},
                    %
                    % the space between legend image and text:
                    /tikz/every odd column/.append style={column sep=0cm},
                },
    ]

\definecolor{c1}{RGB}{141,211,199}
\definecolor{c2}{RGB}{255,255,179}
\definecolor{c3}{RGB}{190,186,218}
\definecolor{c4}{RGB}{251,128,114}
%\definecolor{c5}{RGB}{85,87,179}
\definecolor{c6}{RGB}{253,180,98}

\addplot[
    fill=c1,
    draw=black,
    point meta=y,
    every node near coord/.style={inner ysep=5pt},
    error bars/.cd,
        y dir=both,
        y explicit
] 
table [y] {
x      y  
6   31
8   3
}; \addlegendentry{ SoftML-noiseless\,\,}

\addplot[
    fill=c1,
    postaction={
        pattern=my crosshatch dots
    },
    draw=black,
    point meta=y,
    every node near coord/.style={inner ysep=5pt},
    error bars/.cd,
        y dir=both,
        y explicit
] 
table [y] {
x      y  
6   8
8   0
}; \addlegendentry{ SoftML-with noise\,\,\,\,}

\addplot[
    fill=c2,
    draw=black,
    point meta=y,
    every node near coord/.style={inner ysep=5pt},
    error bars/.cd,
        y dir=both,
        y explicit
] 
table [y] {
x      y     
6   30
8   6   
}; \addlegendentry{ ML-noiseless\,\,}

\addplot[
    fill=c2,
    postaction={
        pattern=my crosshatch dots
    },
    draw=black,
    point meta=y,
    every node near coord/.style={inner ysep=5pt},
    error bars/.cd,
        y dir=both,
        y explicit
] 
table [y] {
x      y     
6   8   
8   0
};\addlegendentry{ ML-with noise\,\,\,\,}

\addplot[
    fill=c4,
    draw=black,
] 
table [y] {
x       y  
8.5   99
}; \addlegendentry{ CNN-noiseless\,\,}

\addplot[
    fill=c4,
    postaction={
        pattern=my crosshatch dots
    },
    draw=black,
] 
table [y] {
x       y  
8.5   97
};\addlegendentry{ CNN-with noise}

\draw [decorate,decoration={brace,amplitude=10pt,raise=4pt},yshift=0pt]
(6.38,-0.12) -- (4.88,-0.12) node [black,midway,yshift=-0.7cm] {
$98\%$};

\draw [decorate,decoration={brace,amplitude=10pt,raise=4pt},yshift=0pt]
(8.39,-0.12) -- (6.88,-0.12) node [black,midway,yshift=-0.7cm] {
$95\%$};

\draw[line width=0.3 mm] (0,0) -- (12,0);
\draw[line width=0.3 mm] (0,100) -- (12,100);
\draw[line width=0.3 mm] (4.791,100) -- (4.791,0);
\draw[line width=0.3 mm] (9.715,100) -- (9.715,0);

%\draw ({rel axis cs:0,0}|-{axis cs:0,0}) -- ({rel axis cs:1,0}|-{axis cs:0,0});
\end{axis}

\begin{pgfonlayer}{bg}    % select the background layer
\begin{axis}[scale=1,
%axis lines*=left,
x axis line style={draw opacity=0},
width=6.1cm,
height=3.4in,
bar width = .3cm,
enlarge x limits=0.49,
    tick align=inside,
    tick style={draw=none},
    xtick={6,8,10},
    xticklabels={%
        \small{$98$},
        \small{$95$},
        \small{CNN}},
     xticklabel style = { font=\scriptsize, text width=2.2cm, align=center },
    ybar=1,
    ymin = -18,
    xmajorticks=false,
    ymax = 120,
ytick={0,  20,  40,  60, 80, 100},
ylabel={ },
    legend image code/.code={%
                    \draw[#1, draw=none] (0cm,-0.1cm) rectangle (0.4cm,0.17cm);
                },  
                legend style={
                    %draw=none, % ?
                    text depth=0pt,
                    at={(0.37,0.99)},
                    anchor=north west,
                    legend columns=2,
                    default spacing:
                    %column sep=1cm,
                    % The text "Legend:"
                    /tikz/column 2/.style={column sep=0pt,font=\bfseries},
                    %
                    % the space between legend image and text:
                    /tikz/every odd column/.append style={column sep=0cm},
                },
    ]

\definecolor{c1}{RGB}{141,211,199}
\definecolor{c2}{RGB}{255,255,179}
\definecolor{c3}{RGB}{190,186,218}
\definecolor{c4}{RGB}{251,128,114}
%\definecolor{c5}{RGB}{85,87,179}
\definecolor{c6}{RGB}{253,180,98}

\addplot[
    pattern=north east lines,
    draw=black,
    point meta=y,
    every node near coord/.style={inner ysep=5pt},
    error bars/.cd,
        y dir=both,
        y explicit
] 
table [y] {
x      y  
6   100
8   100
};

\addplot[
    pattern=north east lines,
    draw=black,
    point meta=y,
    every node near coord/.style={inner ysep=5pt},
    error bars/.cd,
        y dir=both,
        y explicit
] 
table [y] {
x      y  
6   99
8   100
};

\addplot[
    pattern=north east lines,
    draw=black,
    point meta=y,
    every node near coord/.style={inner ysep=5pt},
    error bars/.cd,
        y dir=both,
        y explicit
] 
table [y] {
x      y     
6   100
8   100
};

\addplot[
    pattern=north east lines,
    draw=black,
    point meta=y,
    every node near coord/.style={inner ysep=5pt},
    error bars/.cd,
        y dir=both,
        y explicit
] 
table [y] {
x      y     
6   100
8   100
};

\addplot[
    fill=c4,
    draw=black,
] 
table [y] {
x       y  
8.5   0
};

\addplot[
    fill=c4,
    postaction={
        pattern=my crosshatch dots
    },
    draw=black,
] 
table [y] {
x       y  
8.5   0
};

\draw [decorate,decoration={brace,amplitude=10pt,raise=4pt},yshift=0pt]
(6.38,-0.12) -- (4.88,-0.12) node [black,midway,yshift=-0.7cm] {
$98\%$};

\draw [decorate,decoration={brace,amplitude=10pt,raise=4pt},yshift=0pt]
(8.39,-0.12) -- (6.88,-0.12) node [black,midway,yshift=-0.7cm] {
$95\%$};

\draw[line width=0.3 mm] (0,0) -- (12,0);
\draw[line width=0.3 mm] (0,100) -- (12,100);
\draw[line width=0.3 mm] (4.791,100) -- (4.791,0);
\draw[line width=0.3 mm] (9.715,100) -- (9.715,0);

%\draw ({rel axis cs:0,0}|-{axis cs:0,0}) -- ({rel axis cs:1,0}|-{axis cs:0,0});
\end{axis}
\end{pgfonlayer}

\end{tikzpicture} }
  \subcaption{{\small FGSM}}
  \label{fig:UntargettedFGSM}
\end{minipage}
\begin{minipage}{.24\textwidth}
  \centering
\scalebox{0.72}{\begin{tikzpicture}
\begin{axis}[scale=1,
%axis lines*=left,
x axis line style={draw opacity=0},
width=6.1cm,
height=3.4in,
bar width = .3cm,
enlarge x limits=0.49,
    tick align=inside,
    tick style={draw=none},
    xtick={6,8,10},
    xticklabels={%
        \small{$98$},
        \small{$95$},
        \small{CNN}},
     xticklabel style = { font=\scriptsize, text width=2.2cm, align=center },
    ybar=1,
    ymin = -18,
    xmajorticks=false,
    ymax = 120,
ytick={0,  20,  40,  60, 80, 100},
ylabel={ },
    legend image code/.code={%
                    \draw[#1, draw=none] (0cm,-0.1cm) rectangle (0.4cm,0.17cm);
                },  
                legend style={
                    %draw=none, % ?
                    text depth=0pt,
                    at={(0.37,0.99)},
                    anchor=north west,
                    legend columns=2,
                    default spacing:
                    %column sep=1cm,
                    % The text "Legend:"
                    /tikz/column 2/.style={column sep=0pt,font=\bfseries},
                    %
                    % the space between legend image and text:
                    /tikz/every odd column/.append style={column sep=0cm},
                },
    ]

\definecolor{c1}{RGB}{141,211,199}
\definecolor{c2}{RGB}{255,255,179}
\definecolor{c3}{RGB}{190,186,218}
\definecolor{c4}{RGB}{251,128,114}
%\definecolor{c5}{RGB}{85,87,179}
\definecolor{c6}{RGB}{253,180,98}

\addplot[
    fill=c1,
    draw=black,
    point meta=y,
    every node near coord/.style={inner ysep=5pt},
    error bars/.cd,
        y dir=both,
        y explicit
] 
table [y] {
x      y  
6   18
8   0
};

\addplot[
    fill=c1,
    postaction={
        pattern=my crosshatch dots
    },
    draw=black,
    point meta=y,
    every node near coord/.style={inner ysep=5pt},
    error bars/.cd,
        y dir=both,
        y explicit
] 
table [y] {
x      y  
6   11
8   0
};

\addplot[
    fill=c2,
    draw=black,
    point meta=y,
    every node near coord/.style={inner ysep=5pt},
    error bars/.cd,
        y dir=both,
        y explicit
] 
table [y] {
x      y     
6   14
8   1
};

\addplot[
    fill=c2,
    postaction={
        pattern=my crosshatch dots
    },
    draw=black,
    point meta=y,
    every node near coord/.style={inner ysep=5pt},
    error bars/.cd,
        y dir=both,
        y explicit
] 
table [y] {
x      y     
6   15
8   1
}; 

\addplot[
    fill=c4,
    draw=black,
] 
table [y] {
x       y  
8.5   100
};

\addplot[
    fill=c4,
    postaction={
        pattern=my crosshatch dots
    },
    draw=black,
] 
table [y] {
x       y  
8.5   100
};

\draw [decorate,decoration={brace,amplitude=10pt,raise=4pt},yshift=0pt]
(6.38,-0.12) -- (4.88,-0.12) node [black,midway,yshift=-0.7cm] {
$98\%$};

\draw [decorate,decoration={brace,amplitude=10pt,raise=4pt},yshift=0pt]
(8.39,-0.12) -- (6.88,-0.12) node [black,midway,yshift=-0.7cm] {
$95\%$};

\draw[line width=0.3 mm] (0,0) -- (12,0);
\draw[line width=0.3 mm] (0,100) -- (12,100);
\draw[line width=0.3 mm] (4.791,100) -- (4.791,0);
\draw[line width=0.3 mm] (9.715,100) -- (9.715,0);

%\draw ({rel axis cs:0,0}|-{axis cs:0,0}) -- ({rel axis cs:1,0}|-{axis cs:0,0});
\end{axis}

\begin{pgfonlayer}{bg}    % select the background layer
\begin{axis}[scale=1,
%axis lines*=left,
x axis line style={draw opacity=0},
width=6.1cm,
height=3.4in,
bar width = .3cm,
enlarge x limits=0.49,
    tick align=inside,
    tick style={draw=none},
    xtick={6,8,10},
    xticklabels={%
        \small{$98$},
        \small{$95$},
        \small{CNN}},
     xticklabel style = { font=\scriptsize, text width=2.2cm, align=center },
    ybar=1,
    ymin = -18,
    xmajorticks=false,
    ymax = 120,
ytick={0,  20,  40,  60, 80, 100},
ylabel={ },
    legend image code/.code={%
                    \draw[#1, draw=none] (0cm,-0.1cm) rectangle (0.4cm,0.17cm);
                },  
                legend style={
                    %draw=none, % ?
                    text depth=0pt,
                    at={(0.37,0.99)},
                    anchor=north west,
                    legend columns=2,
                    default spacing:
                    %column sep=1cm,
                    % The text "Legend:"
                    /tikz/column 2/.style={column sep=0pt,font=\bfseries},
                    %
                    % the space between legend image and text:
                    /tikz/every odd column/.append style={column sep=0cm},
                },
    ]

\definecolor{c1}{RGB}{141,211,199}
\definecolor{c2}{RGB}{255,255,179}
\definecolor{c3}{RGB}{190,186,218}
\definecolor{c4}{RGB}{251,128,114}
%\definecolor{c5}{RGB}{85,87,179}
\definecolor{c6}{RGB}{253,180,98}

\addplot[
    pattern=north east lines,
    draw=black,
    point meta=y,
    every node near coord/.style={inner ysep=5pt},
    error bars/.cd,
        y dir=both,
        y explicit
] 
table [y] {
x      y  
6   100
8   100
};

\addplot[
    pattern=north east lines,
    draw=black,
    point meta=y,
    every node near coord/.style={inner ysep=5pt},
    error bars/.cd,
        y dir=both,
        y explicit
] 
table [y] {
x      y  
6   100
8   100
};

\addplot[
    pattern=north east lines,
    draw=black,
    point meta=y,
    every node near coord/.style={inner ysep=5pt},
    error bars/.cd,
        y dir=both,
        y explicit
] 
table [y] {
x      y     
6   100
8   100
};

\addplot[
    pattern=north east lines,
    draw=black,
    point meta=y,
    every node near coord/.style={inner ysep=5pt},
    error bars/.cd,
        y dir=both,
        y explicit
] 
table [y] {
x      y     
6   99
8   100
};

\addplot[
    fill=c4,
    draw=black,
] 
table [y] {
x       y  
8.5   0
};

\addplot[
    fill=c4,
    postaction={
        pattern=my crosshatch dots
    },
    draw=black,
] 
table [y] {
x       y  
8.5   0
};

\draw [decorate,decoration={brace,amplitude=10pt,raise=4pt},yshift=0pt]
(6.38,-0.12) -- (4.88,-0.12) node [black,midway,yshift=-0.7cm] {
$98\%$};

\draw [decorate,decoration={brace,amplitude=10pt,raise=4pt},yshift=0pt]
(8.39,-0.12) -- (6.88,-0.12) node [black,midway,yshift=-0.7cm] {
$95\%$};

\draw[line width=0.3 mm] (0,0) -- (12,0);
\draw[line width=0.3 mm] (0,100) -- (12,100);
\draw[line width=0.3 mm] (4.791,100) -- (4.791,0);
\draw[line width=0.3 mm] (9.715,100) -- (9.715,0);

%\draw ({rel axis cs:0,0}|-{axis cs:0,0}) -- ({rel axis cs:1,0}|-{axis cs:0,0});
\end{axis}
\end{pgfonlayer}

\end{tikzpicture} }
  \subcaption{{\small IGSA}}
  \label{fig:UntargettedIGSA}
\end{minipage}%
\begin{minipage}{.24\textwidth}
  \centering
\scalebox{0.72}{\begin{tikzpicture}
\begin{axis}[scale=1,
%axis lines*=left,
x axis line style={draw opacity=0},
width=6.1cm,
height=3.4in,
bar width = .3cm,
enlarge x limits=0.49,
    tick align=inside,
    tick style={draw=none},
    xtick={6,8,10},
    xticklabels={%
        \small{$98$},
        \small{$95$},
        \small{CNN}},
     xticklabel style = { font=\scriptsize, text width=2.2cm, align=center },
    ybar=1,
    ymin = -18,
    xmajorticks=false,
    ymax = 120,
ytick={0,  20,  40,  60, 80, 100},
ylabel={ },
    legend image code/.code={%
                    \draw[#1, draw=none] (0cm,-0.1cm) rectangle (0.4cm,0.17cm);
                },  
                legend style={
                    %draw=none, % ?
                    text depth=0pt,
                    at={(0.41,0.99)},
                    anchor=north west,
                    legend columns=2,
                    default spacing:
                    %column sep=1cm,
                    % The text "Legend:"
                    /tikz/column 2/.style={column sep=0pt,font=\bfseries},
                    %
                    % the space between legend image and text:
                    /tikz/every odd column/.append style={column sep=0cm},
                },
    ]

\definecolor{c1}{RGB}{141,211,199}
\definecolor{c2}{RGB}{255,255,179}
\definecolor{c3}{RGB}{190,186,218}
\definecolor{c4}{RGB}{251,128,114}
%\definecolor{c5}{RGB}{85,87,179}
\definecolor{c6}{RGB}{253,180,98}

\addplot[
    fill=c1,
    draw=black,
    point meta=y,
    every node near coord/.style={inner ysep=5pt},
    error bars/.cd,
        y dir=both,
        y explicit
] 
table [y] {
x      y  
6   25
8   3
};

\addplot[
    fill=c1,
    postaction={
        pattern=my crosshatch dots
    },
    draw=black,
    point meta=y,
    every node near coord/.style={inner ysep=5pt},
    error bars/.cd,
        y dir=both,
        y explicit
] 
table [y] {
x      y  
6   23
8   1
};

\addplot[
    fill=c2,
    draw=black,
    point meta=y,
    every node near coord/.style={inner ysep=5pt},
    error bars/.cd,
        y dir=both,
        y explicit
] 
table [y] {
x      y     
6   24
8   6
}; 

\addplot[
    fill=c2,
    postaction={
        pattern=my crosshatch dots
    },
    draw=black,
    point meta=y,
    every node near coord/.style={inner ysep=5pt},
    error bars/.cd,
        y dir=both,
        y explicit
] 
table [y] {
x      y     
6   25   
8   1
};

\addplot[
    fill=c4,
    draw=black,
] 
table [y] {
x       y  
8.5   47
};

\addplot[
    fill=c4,
    postaction={
        pattern=my crosshatch dots
    },
    draw=black,
] 
table [y] {
x       y  
8.5   76
};

\draw [decorate,decoration={brace,amplitude=10pt,raise=4pt},yshift=0pt]
(6.38,-0.12) -- (4.88,-0.12) node [black,midway,yshift=-0.7cm] {
$98\%$};

\draw [decorate,decoration={brace,amplitude=10pt,raise=4pt},yshift=0pt]
(8.39,-0.12) -- (6.88,-0.12) node [black,midway,yshift=-0.7cm] {
$95\%$};

\draw[line width=0.3 mm] (0,0) -- (12,0);
\draw[line width=0.3 mm] (0,100) -- (12,100);
\draw[line width=0.3 mm] (4.791,100) -- (4.791,0);
\draw[line width=0.3 mm] (9.715,100) -- (9.715,0);

%\draw ({rel axis cs:0,0}|-{axis cs:0,0}) -- ({rel axis cs:1,0}|-{axis cs:0,0});
\end{axis}

\begin{pgfonlayer}{bg}    % select the background layer
\begin{axis}[scale=1,
%axis lines*=left,
x axis line style={draw opacity=0},
width=6.1cm,
height=3.4in,
bar width = .3cm,
enlarge x limits=0.49,
    tick align=inside,
    tick style={draw=none},
    xtick={6,8,10},
    xticklabels={%
        \small{$98$},
        \small{$95$},
        \small{CNN}},
     xticklabel style = { font=\scriptsize, text width=2.2cm, align=center },
    ybar=1,
    ymin = -18,
    xmajorticks=false,
    ymax = 120,
ytick={0,  20,  40,  60, 80, 100},
ylabel={ },
    legend image code/.code={%
                    \draw[#1, draw=none] (0cm,-0.1cm) rectangle (0.4cm,0.17cm);
                },  
                legend style={
                    %draw=none, % ?
                    text depth=0pt,
                    at={(0.37,0.99)},
                    anchor=north west,
                    legend columns=1,
                    default spacing:
                    %column sep=1cm,
                    % The text "Legend:"
                    /tikz/column 2/.style={column sep=0pt,font=\bfseries},
                    %
                    % the space between legend image and text:
                    /tikz/every odd column/.append style={column sep=0cm},
                },
    ]

\definecolor{c1}{RGB}{141,211,199}
\definecolor{c2}{RGB}{255,255,179}
\definecolor{c3}{RGB}{190,186,218}
\definecolor{c4}{RGB}{251,128,114}
%\definecolor{c5}{RGB}{85,87,179}
\definecolor{c6}{RGB}{253,180,98}

\addplot[
    pattern=north east lines,
    draw=black,
    point meta=y,
    every node near coord/.style={inner ysep=5pt},
    error bars/.cd,
        y dir=both,
        y explicit
] 
table [y] {
x      y  
6   99
8   100
}; 

\addplot[
    pattern=north east lines,
    draw=black,
    point meta=y,
    every node near coord/.style={inner ysep=5pt},
    error bars/.cd,
        y dir=both,
        y explicit
] 
table [y] {
x      y  
6   92
8   99
};

\addplot[
    pattern=north east lines,
    draw=black,
    point meta=y,
    every node near coord/.style={inner ysep=5pt},
    error bars/.cd,
        y dir=both,
        y explicit
] 
table [y] {
x      y     
6   97
8   100
};

\addplot[
    pattern=north east lines,
    draw=black,
    point meta=y,
    every node near coord/.style={inner ysep=5pt},
    error bars/.cd,
        y dir=both,
        y explicit
] 
table [y] {
x      y     
6   89
8   97
};

\addplot[
    fill=c4,
    draw=black,
] 
table [y] {
x       y  
8.5   0
};

\addplot[
    fill=c4,
    postaction={
        pattern=my crosshatch dots
    },
    draw=black,
] 
table [y] {
x       y  
8.5   0
};

\draw [decorate,decoration={brace,amplitude=10pt,raise=4pt},yshift=0pt]
(6.38,-0.12) -- (4.88,-0.12) node [black,midway,yshift=-0.7cm] {
$98\%$};

\draw [decorate,decoration={brace,amplitude=10pt,raise=4pt},yshift=0pt]
(8.39,-0.12) -- (6.88,-0.12) node [black,midway,yshift=-0.7cm] {
$95\%$};

\draw[line width=0.3 mm] (0,0) -- (12,0);
\draw[line width=0.3 mm] (0,100) -- (12,100);
\draw[line width=0.3 mm] (4.791,100) -- (4.791,0);
\draw[line width=0.3 mm] (9.715,100) -- (9.715,0);

%\draw ({rel axis cs:0,0}|-{axis cs:0,0}) -- ({rel axis cs:1,0}|-{axis cs:0,0});
\end{axis}
\end{pgfonlayer}

\end{tikzpicture} }
  \subcaption{{\small GA}}
  \label{fig:UntargettedGA}
\end{minipage}
\begin{minipage}{.24\textwidth}
  \centering
\scalebox{0.72}{\begin{tikzpicture}
\begin{axis}[scale=1,
%axis lines*=left,
x axis line style={draw opacity=0},
width=6.1cm,
height=3.4in,
bar width = .3cm,
enlarge x limits=0.49,
    tick align=inside,
    tick style={draw=none},
    xtick={6,8,10},
    xticklabels={%
        \small{$98$},
        \small{$95$},
        \small{CNN}},
     xticklabel style = { font=\scriptsize, text width=2.2cm, align=center },
    ybar=1,
    ymin = -18,
    xmajorticks=false,
    ymax = 120,
ytick={0,  20,  40,  60, 80, 100},
ylabel={ },
    legend image code/.code={%
                    \draw[#1, draw=none] (0cm,-0.1cm) rectangle (0.4cm,0.17cm);
                },  
                legend style={
                    %draw=none, % ?
                    text depth=0pt,
                    at={(0.42,0.99)},
                    anchor=north west,
                    legend columns=2,
                    default spacing:
                    %column sep=1cm,
                    % The text "Legend:"
                    /tikz/column 2/.style={column sep=0pt,font=\bfseries},
                    %
                    % the space between legend image and text:
                    /tikz/every odd column/.append style={column sep=0cm},
                },
    ]

\definecolor{c1}{RGB}{141,211,199}
\definecolor{c2}{RGB}{255,255,179}
\definecolor{c3}{RGB}{190,186,218}
\definecolor{c4}{RGB}{251,128,114}
%\definecolor{c5}{RGB}{85,87,179}
\definecolor{c6}{RGB}{253,180,98}

\addplot[
    fill=c1,
    draw=black,
    point meta=y,
    every node near coord/.style={inner ysep=5pt},
    error bars/.cd,
        y dir=both,
        y explicit
] 
table [y] {
x      y  
6   25
8   2
};

\addplot[
    fill=c1,
    postaction={
        pattern=my crosshatch dots
    },
    draw=black,
    point meta=y,
    every node near coord/.style={inner ysep=5pt},
    error bars/.cd,
        y dir=both,
        y explicit
] 
table [y] {
x      y  
6   26
8   1
};

\addplot[
    fill=c2,
    draw=black,
    point meta=y,
    every node near coord/.style={inner ysep=5pt},
    error bars/.cd,
        y dir=both,
        y explicit
] 
table [y] {
x      y     
6   24
8   4   
};

\addplot[
    fill=c2,
    postaction={
        pattern=my crosshatch dots
    },
    draw=black,
    point meta=y,
    every node near coord/.style={inner ysep=5pt},
    error bars/.cd,
        y dir=both,
        y explicit
] 
table [y] {
x      y     
6   27
8   1
};

\addplot[
    fill=c4,
    draw=black,
] 
table [y] {
x       y  
8.5   100
};

\addplot[
    fill=c4,
    postaction={
        pattern=my crosshatch dots
    },
    draw=black,
] 
table [y] {
x       y  
8.5   100
};

\draw [decorate,decoration={brace,amplitude=10pt,raise=4pt},yshift=0pt]
(6.38,-0.12) -- (4.88,-0.12) node [black,midway,yshift=-0.7cm] {
$98\%$};

\draw [decorate,decoration={brace,amplitude=10pt,raise=4pt},yshift=0pt]
(8.39,-0.12) -- (6.88,-0.12) node [black,midway,yshift=-0.7cm] {
$95\%$};

\draw[line width=0.3 mm] (0,0) -- (12,0);
\draw[line width=0.3 mm] (0,100) -- (12,100);
\draw[line width=0.3 mm] (4.791,100) -- (4.791,0);
\draw[line width=0.3 mm] (9.715,100) -- (9.715,0);

%\draw ({rel axis cs:0,0}|-{axis cs:0,0}) -- ({rel axis cs:1,0}|-{axis cs:0,0});
\end{axis}

\begin{pgfonlayer}{bg}    % select the background layer
\begin{axis}[scale=1,
%axis lines*=left,
x axis line style={draw opacity=0},
width=6.1cm,
height=3.4in,
bar width = .3cm,
enlarge x limits=0.49,
    tick align=inside,
    tick style={draw=none},
    xtick={6,8,10},
    xticklabels={%
        \small{$98$},
        \small{$95$},
        \small{CNN}},
     xticklabel style = { font=\scriptsize, text width=2.2cm, align=center },
    ybar=1,
    ymin = -18,
    xmajorticks=false,
    ymax = 120,
ytick={0,  20,  40,  60, 80, 100},
ylabel={ },
    legend image code/.code={%
                    \draw[#1, draw] (0cm,-0.1cm) rectangle (0.4cm,0.17cm);
                },  
                legend style={
                    %draw=none, % ?
                    text depth=0pt,
                    at={(0.53,0.99)},
                    anchor=north west,
                    legend columns=1,
                    default spacing:
                    %column sep=1cm,
                    % The text "Legend:"
                    /tikz/column 2/.style={column sep=0pt,font=\bfseries},
                    %
                    % the space between legend image and text:
                    /tikz/every odd column/.append style={column sep=0cm},
                },
    ]

\definecolor{c1}{RGB}{141,211,199}
\definecolor{c2}{RGB}{255,255,179}
\definecolor{c3}{RGB}{190,186,218}
\definecolor{c4}{RGB}{251,128,114}
%\definecolor{c5}{RGB}{85,87,179}
\definecolor{c6}{RGB}{253,180,98}

\addplot[
    pattern=north east lines,
    draw=black,
    point meta=y,
    every node near coord/.style={inner ysep=5pt},
    error bars/.cd,
        y dir=both,
        y explicit
] 
table [y] {
x      y  
6   100
8   100
}; \addlegendentry{ Rejected}

\addplot[
    pattern=north east lines,
    draw=black,
    point meta=y,
    every node near coord/.style={inner ysep=5pt},
    error bars/.cd,
        y dir=both,
        y explicit
] 
table [y] {
x      y  
6   100
8   100
};

\addplot[
    pattern=north east lines,
    draw=black,
    point meta=y,
    every node near coord/.style={inner ysep=5pt},
    error bars/.cd,
        y dir=both,
        y explicit
] 
table [y] {
x      y     
6   100
8   100
};

\addplot[
    pattern=north east lines,
    draw=black,
    point meta=y,
    every node near coord/.style={inner ysep=5pt},
    error bars/.cd,
        y dir=both,
        y explicit
] 
table [y] {
x      y     
6   100
8   100
};

\addplot[
    fill=c4,
    draw=black,
] 
table [y] {
x       y  
8.5   0
};

\addplot[
    fill=c4,
    postaction={
        pattern=my crosshatch dots
    },
    draw=black,
] 
table [y] {
x       y  
8.5   0
};

\draw [decorate,decoration={brace,amplitude=10pt,raise=4pt},yshift=0pt]
(6.38,-0.12) -- (4.88,-0.12) node [black,midway,yshift=-0.7cm] {
$98\%$};

\draw [decorate,decoration={brace,amplitude=10pt,raise=4pt},yshift=0pt]
(8.39,-0.12) -- (6.88,-0.12) node [black,midway,yshift=-0.7cm] {
$95\%$};

\draw[line width=0.3 mm] (0,0) -- (12,0);
\draw[line width=0.3 mm] (0,100) -- (12,100);
\draw[line width=0.3 mm] (4.791,100) -- (4.791,0);
\draw[line width=0.3 mm] (9.715,100) -- (9.715,0);

%\draw ({rel axis cs:0,0}|-{axis cs:0,0}) -- ({rel axis cs:1,0}|-{axis cs:0,0});
\end{axis}
\end{pgfonlayer}

\end{tikzpicture} }
  \subcaption{{\small L-BFGS}}
  \label{fig:UntargettedL-BFGS}
\end{minipage}
\caption{{\small The success rate of the untargeted adversarial attacks FGSM, IGSA, GA, and L-BFGS against the deep-RBF network and CNN. Here, we try to decrease the score of the correct class. The attack is only considered successful when the found distorted example has a non-negative label. The hatched sections show the percentage of the adversarial examples with a negative label, i.e., the deep-RBF network rejects these adversarial examples. We can see that most distorted examples adversarial attacks found for the deep-RBF networks have the negative class label, which is not harmful.}}
\label{fig:ADVsUntargeted}
\end{figure}
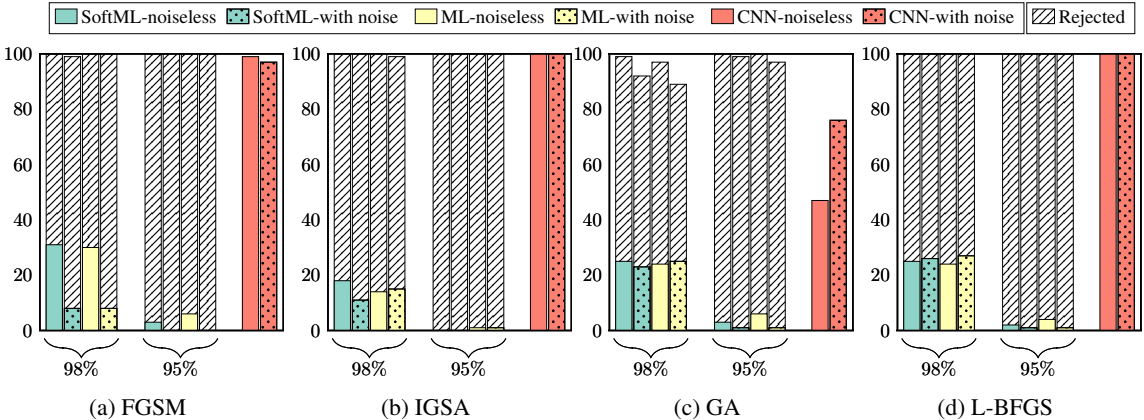

Thus far, we have evaluated the resistance of the deep-RBF network only to the targeted attacks; we now want to analyze our method when the attacks are untargeted. The aim of this attack is decreasing the score (confidence) of the correct class so that misclassification occurs. Fig.~\ref{fig:ADVsUntargeted} shows the success rate of the attacks FGSM, IGSA, GA and L-BFGS on the deep-RBF network, together with the ordinary CNN. As we can see, the results of our methods are pretty much the same, and the attacks often result in a distorted example whose class is negative. Therefore, the sample will be rejected, and our methods will not be fooled. That is why the success rate for our methods is often small especially when the threshold is low (the case with $95\%$ accuracy). Here, like the targeted version, the version with noise injection perform better than the noiseless one.

\subsubsection{Targeted L-BFGS attack}
At the end of this section, we want to test the resistance of the deep-RBF network to the targeted L-BFGS attack~\citep{szegedy2013intriguing}. The results for its untargeted version appear in Fig.~\ref{fig:ADVsUntargeted}. This attack is more potent than FGSM, IGSA, and GA; it can often find an adversarial example in the targeted version. However, we claim that when this attack is performed on our methods, it usually distorts an image such that its correct class also changes. In these cases, we can say that the attack moves on the manifold of the dataset. If we call it fooling, sometimes, it can even fools humans. In Table~\ref{table:L-BFGS}, we report some measures of the found adversarial examples for two cases: ordinary and rubbish. In the ordinary case, the input samples are the first 100 images of the MNIST test dataset whereas in the rubbish case, the input samples are 200 isotropic Gaussian noises. The mean of the noise is $0.5$, and its standard deviation is $0.1$ while each pixel takes a value between zero and one. Here, MAD is the average of Mean Absolute distortion, and L-inf expresses the maximum of the differences between the pixels of the original image and the distorted one (i.e., the L-infinity distance). We have not mentioned the ratio of the average confidences (RoC) for the rubbish case because the rubbish examples do not have a valid class.

\begin{table}
\scriptsize
\caption{\small The targeted adversarial attack L-BFGS on the deep-RBF networks and the CNN. In the ordinary case, the input samples are the first 100 images of the MNIST test dataset, and in the rubbish case, the input samples are 200 isotropic Gaussian noises. MAD is the average of Mean Absolute distortion, and L-inf is the L-infinity distance between the original and distorted images.}
\label{table:L-BFGS}
\begin{center}
\begin{footnotesize}
\begin{tabular}{c|c|c|c|c|||c|c|c|}
\cline{2-8}
   & \multicolumn{4}{c|||}{{\bf Ordinary}}  & \multicolumn{3}{c|}{{\bf Rubbish}} \\ \hline 
\multicolumn{1}{|c||}{{\bf Methods}} &    {\bf  RoC}     &    {\bf RMSD}    &    {\bf MAD}    &     {\bf L-inf}     &    {\bf RMSD}    &    {\bf MAD}    &     {\bf L-inf}   \\ \hline \hline
\multicolumn{1}{|c||}{{\bf SoftML loss} -- noiseless -- 98\%\,\,\,}	     &     4e+39    &   0.073     &    0.049     &  0.347    &   0.063     &    0.048    &   0.259   \\ \hline
\multicolumn{1}{|c||}{{\bf SoftML loss} -- with noise -- 98\%}        &    3e+11     &    0.111    &    0.059     &   0.727    &    0.075    &     0.054     &  0.325    \\ \hline \hline
\multicolumn{1}{|c||}{{\bf ML loss} -- noiseless -- 98\%\,\,\,}	     &     5e+13    &   0.072     &    0.047     &  0.339    &   0.040     &    0.030    &   0.183   \\ \hline
\multicolumn{1}{|c||}{{\bf ML loss} -- with noise -- 98\%}        &    8.5e+5     &    0.116    &    0.059     &   0.737    &    0.089    &     0.064     &  0.389    \\ \hline \hline \hline
\multicolumn{1}{|c||}{{\bf SoftML loss} -- noiseless -- 95\%\,\,\,}	     &     6.4e+6    &   0.063     &    0.040     &  0.334    &   0.054     &    0.041    &   0.228   \\ \hline
\multicolumn{1}{|c||}{{\bf SoftML loss} -- with noise -- 95\%}        &    5150     &    0.105    &    0.048     &   0.714    &    0.079    &     0.058     &  0.338    \\ \hline \hline 
\multicolumn{1}{|c||}{{\bf ML loss} -- noiseless -- 95\%\,\,\,}	  &     2100    &    0.063    &    0.039    &   0.336     &    0.040    &    0.030     &   0.187     \\ \hline 
\multicolumn{1}{|c||}{{\bf ML loss} -- with noise -- 95\%}          &    12374     &    0.108    &     0.050    &   0.725      &    0.094    &    0.067     &   0.403    \\ \hline \hline \hline
\multicolumn{1}{|c||}{{\bf CNN} -- noiseless\,\,\,}	&     2.061    &    0.049    &    0.027    &   0.305     &    0.019    &     0.010    &  0.125    \\ \hline
\multicolumn{1}{|c||}{{\bf CNN} -- with noise}     &   2.132     &    0.075     &    0.033    &   0.538    &    0.010    &    0.007     &  0.050   \\ \hline
\end{tabular}
\end{footnotesize}
\end{center}
\end{table}

\begin{figure}[]
\centering
        \includegraphics[totalheight=2.5cm]{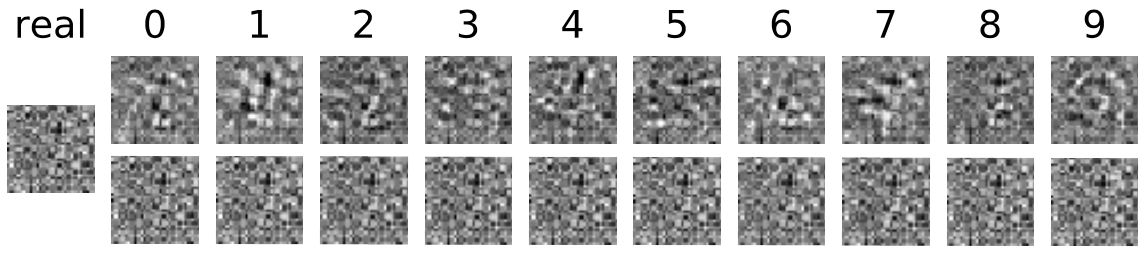}
    \caption{{\small The adversarial examples for the deep-RBF network (the first row) and CNN (the second row), found by the targeted L-BFGS attack. The input sample (real) is a rubbish image which appears in the first column. While the change in the second row is imperceptible, we can easily notice a change in the adversarial examples of the deep-RBF network. Also, if we look carefully, we may even see the pattern of each digit.}}
    \label{fig:rubbish}
\end{figure}

\begin{figure}
\centering
\begin{minipage}{.5\textwidth}
  \centering
  \includegraphics[totalheight=9cm, clip, trim=3cm 0cm 2.5cm 0cm]{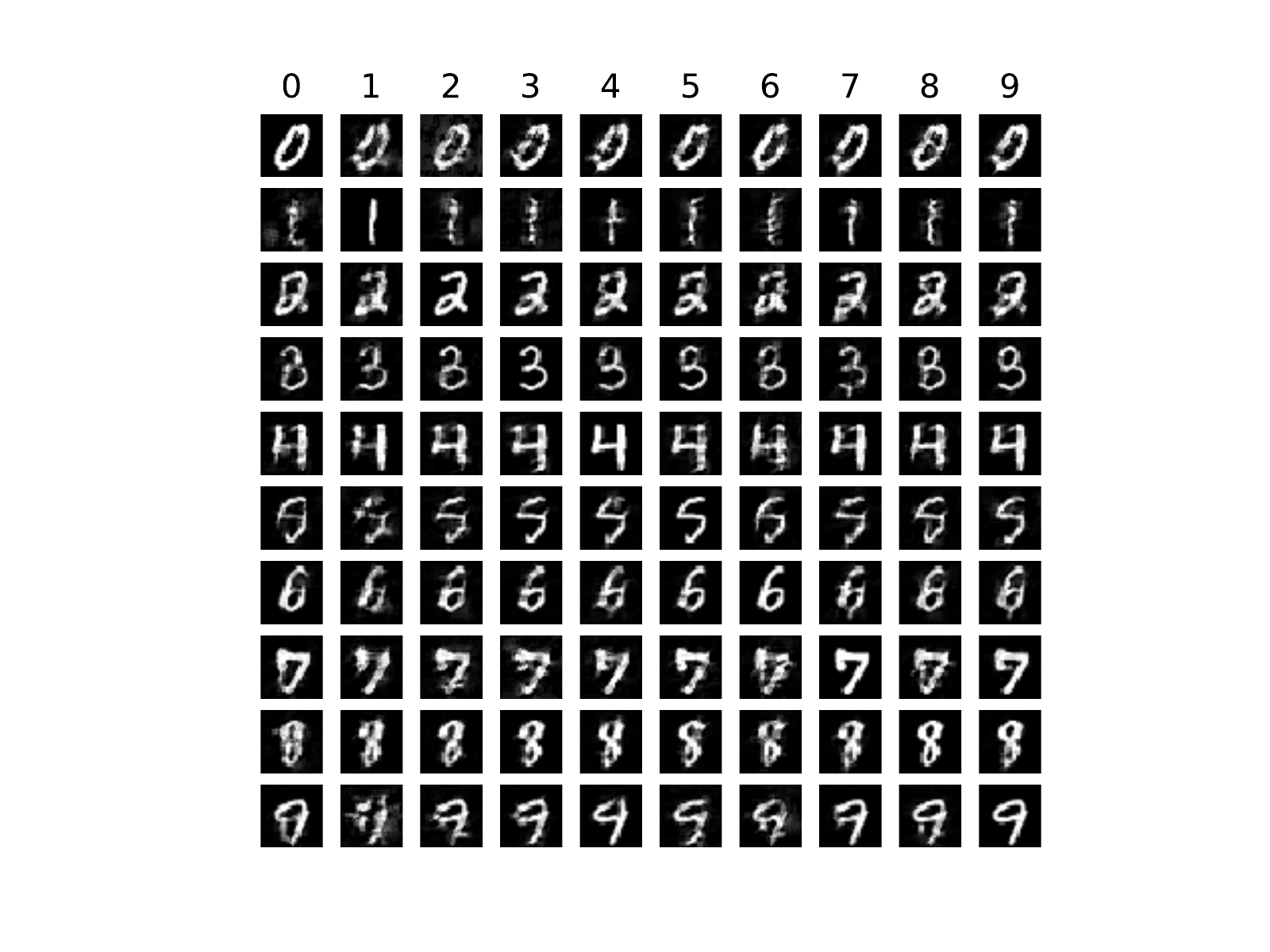}
  \subcaption{{\small Deep-RBF network}}
  \label{fig:L-BFGS_NCC}
\end{minipage}%
\begin{minipage}{.5\textwidth}
  \centering
  \includegraphics[totalheight=9cm, clip, trim=3cm 0cm 2.5cm 0cm]{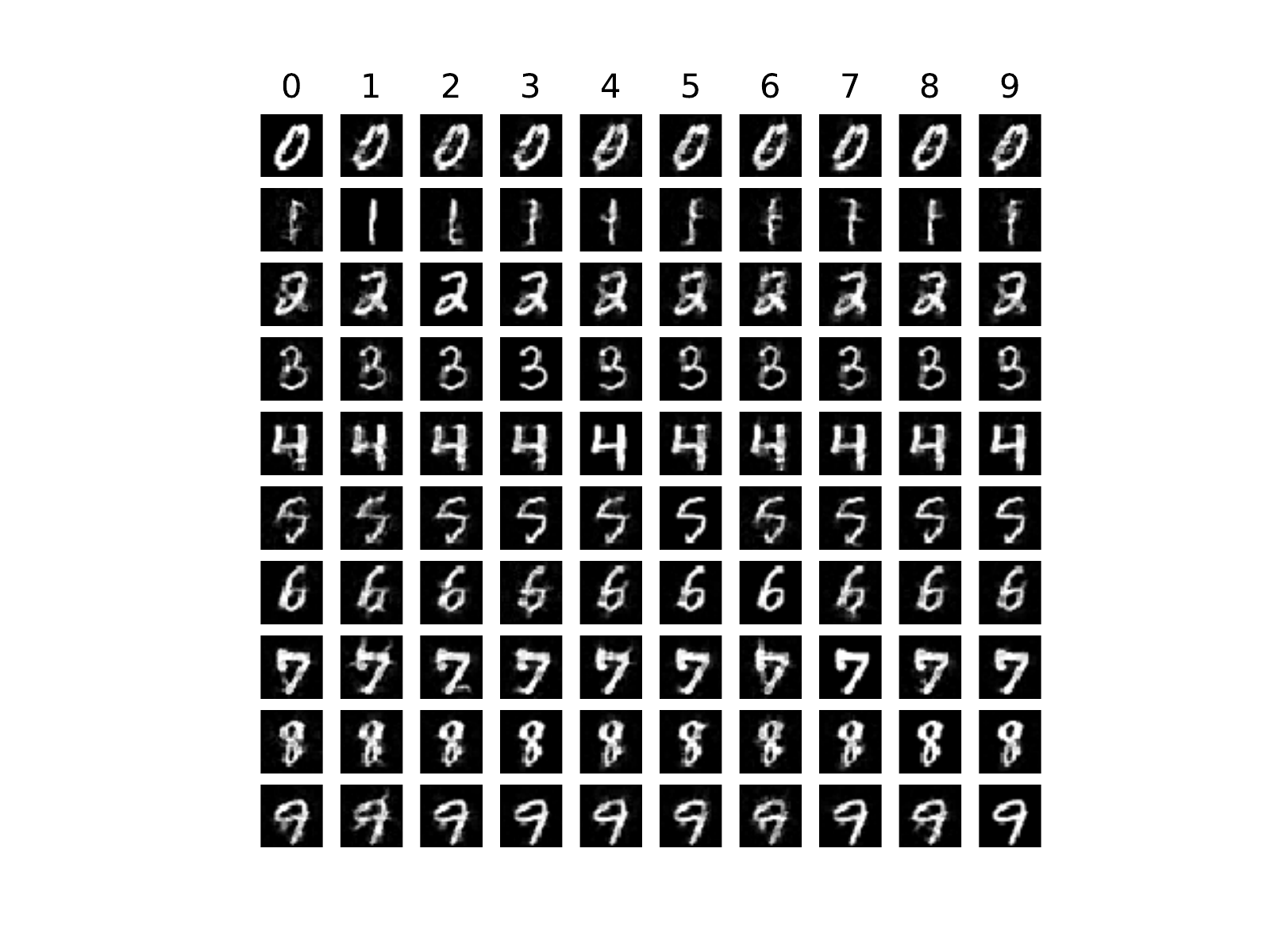}
  \subcaption{{\small CNN}}
  \label{fig:L-BFGS_CNN}
\end{minipage}
\caption{{\small The adversarial examples for the deep-RBF network and the ordinary convolutional neural network found by the targeted L-BFGS attack. The input samples appear in the main diagonal of each plot, and each column is associated with a target for the attacks. The plots on the left (a) and right (b) show the adversarial examples of the networks deep-RBF and CNN respectively. By comparing these plots carefully, we can admit that the magnitude of distortion in the adversarial examples of the deep-RBF network is bigger than CNN. Also, we can see that the adversarial examples of the deep-RBF network are usually more similar to the examples whose classes are the target label, and some of them may even fool humans.}}
\label{fig:L-BFGS}
\end{figure}

The measures in Table~\ref{table:L-BFGS} indicates that our methods, especially the case with noise injection, are more robust than the ordinary CNN because the original image should change more to fool the network. However, these measures are not necessarily perfect criteria because the direction of change in the images does not affect them; the image could just become noisy, blurry, or could switch to another class, and these measures do not distinguish them. Therefore, we draw figures~\ref{fig:rubbish} and~\ref{fig:L-BFGS} to show that the adversarial examples of our methods found by L-BFGS are usually somewhere near the manifold of the target label of the attack. Fig.~\ref{fig:rubbish} shows the adversarial examples of the networks deep-RBF and CNN when noise injection is performed. The attack is targeted L-BFGS, and the input sample (real image) for all attacks is one random sample of Gaussian noise (i.e., a rubbish image). While the first row is the adversarial examples of the deep-RBF network for different targets, the second row is for CNN. We can see that the change in the adversarial images of CNN is imperceptible whereas, in the first row, where the adversarial images of our method exist, we can quickly notice a change. Moreover, if we look carefully, we may see the pattern of each digit in the image with its corresponding target. This image suggests that the direction of change in the adversarial examples of our method is also correct.

Fig.~\ref{fig:L-BFGS} displays the adversarial examples of the deep-RBF network and the CNN. The networks and experiments are the same as Fig.~\ref{fig:rubbish}, but with ten ordinary examples instead of rubbish example. The images in the main diagonal of each plot are the input samples of the attacks, and each column is a target label for the attacks. The plot on the left (a) shows the adversarial examples for deep-RBF, and the examples in the right one (b) are for CNN. Both networks have been trained with noise injection.
If we compare these plots, we can admit that the magnitude of distortion in the adversarial examples of our method is bigger than the magnitude of distortion in those of CNN. Also, if we look at the left plot carefully, we can see that the adversarial examples of the deep-RBF network are usually similar to the examples whose classes are the target label, and they may sometimes actually switch to those classes (and fool humans).
Therefore, something like morphing is happening when the L-BFGS attack runs on a deep-RBF network because the adversarial examples are near, and sometimes on, the manifold of images.
% One might think of our method as a supervised manifold learning.
%\footnote{ The cost function of the targeted L-BFGS tries to minimize the cross-entropy of the distorted image for the target label and the norm of distortion at the same time. We should, like \citep{simonyan2013deep}, maximize only the score of a target and see if it morphs.}.
%In addition to the fact that the adversarial examples L-BFGS attack found for the deep-RBF networks are better than the adversarial examples of CNN, the ratio of the confidences for these examples is much bigger than the RoC of the CNN's adversarial examples. 
Note that in the left plot of this figure, the confidence of the adversarial examples L-BFGS attack found is much smaller than the confidence of the original examples, unlike the right plot.

Table~\ref{table:cross_L-BFGS} reports the results for the cross-model generalization of adversarial examples when the targeted L-BFGS attack is performed. We can see that the percentage of transferable adversarial examples between CNNs is tiny. Furthermore, the percentage of adversarial examples of CNN that deep-RBF network rejects is also relatively small. It means that the distortion of the adversarial examples which are successful in fooling the ordinary CNN is so insignificant that the deep-RBF network would not even reject them. On the other hand, the adversarial examples found for the deep-RBF network with noise injection fool other networks more frequently.

\begin{table}
\setlength{\tabcolsep}{4.5pt}
\scriptsize
\caption{{\small The cross-model generalization of adversarial examples when the targeted L-BFGS attack runs.}}
\label{table:cross_L-BFGS}
\begin{center}
\begin{small}
\begin{tabular}{c|c|c|c|c|c|c|}
\cline{2-7}
                       & {\begin{tabular}[c]{@{}c@{}}{\bf CNN}\\ noiseless\end{tabular}} & {\begin{tabular}[c]{@{}c@{}}{\bf CNN}\\ with noise\end{tabular}} & {\begin{tabular}[c]{@{}c@{}}{\bf Deep-RBF}\\ 98\% - noiseless\end{tabular}} & {\begin{tabular}[c]{@{}c@{}}{\bf Deep-RBF}\\ 98\% - with noise\end{tabular}} & {\begin{tabular}[c]{@{}c@{}}{\bf Deep-RBF}\\ 95\% - noiseless\end{tabular}} &{\begin{tabular}[c]{@{}c@{}}{\bf Deep-RBF}\\ 95\% - with noise\end{tabular}} \\ \hline 
\multicolumn{1}{|c||}{\begin{tabular}[c]{@{}c@{}}{\bf CNN}\\ noiseless\end{tabular}} & \begin{tabular}[c]{@{}c@{}}100\%\\ —\end{tabular} & \begin{tabular}[c]{@{}c@{}}2.22\%\\ —\end{tabular} & \begin{tabular}[c]{@{}c@{}}0.55\%\\ (2.22\%)\end{tabular} & \begin{tabular}[c]{@{}c@{}}0.00\%\\ (3.00\%)\end{tabular} & \begin{tabular}[c]{@{}c@{}}0.11\%\\ (6.55\%)\end{tabular} & \begin{tabular}[c]{@{}c@{}}0.00\%\\ (8.44\%)\end{tabular} \\ \hline
\multicolumn{1}{|c||}{\begin{tabular}[c]{@{}c@{}}{\bf CNN}\\ with noise\end{tabular}} & \begin{tabular}[c]{@{}c@{}}7.66\%\\ —\end{tabular} & \begin{tabular}[c]{@{}c@{}}100\%\\ —\end{tabular} & \begin{tabular}[c]{@{}c@{}}2.88\%\\ (5.44\%)\end{tabular} & \begin{tabular}[c]{@{}c@{}}0.44\%\\ (6.00\%)\end{tabular} & \begin{tabular}[c]{@{}c@{}}1.22\%\\ (24.4\%)\end{tabular} & \begin{tabular}[c]{@{}c@{}}0.00\%\\ (23.3\%)\end{tabular} \\ \hline
\multicolumn{1}{|c||}{\begin{tabular}[c]{@{}c@{}}{\bf Deep-RBF}\\ 98\% - noiseless\end{tabular}} & \begin{tabular}[c]{@{}c@{}}4.03\%\\ —\end{tabular} & \begin{tabular}[c]{@{}c@{}}3.47\%\\ —\end{tabular} & \begin{tabular}[c]{@{}c@{}}100\%\\ (0\%)\end{tabular} & \begin{tabular}[c]{@{}c@{}}0.11\%\\ (6.04\%)\end{tabular} & \begin{tabular}[c]{@{}c@{}}28.4\%\\ (71.5\%)\end{tabular} & \begin{tabular}[c]{@{}c@{}}0.11\%\\ (11.7\%)\end{tabular} \\  \hline
\multicolumn{1}{|c||}{\begin{tabular}[c]{@{}c@{}}{\bf Deep-RBF}\\ 98\% - with noise\end{tabular}} & \begin{tabular}[c]{@{}c@{}}20.49\%\\ —\end{tabular} & \begin{tabular}[c]{@{}c@{}}17.8\%\\ —\end{tabular} & \begin{tabular}[c]{@{}c@{}}26.2\%\\ (18.4\%)\end{tabular} & \begin{tabular}[c]{@{}c@{}}100\%\\ (0\%)\end{tabular} & \begin{tabular}[c]{@{}c@{}}11.5\%\\ (54.3\%)\end{tabular} & \begin{tabular}[c]{@{}c@{}}23.4\%\\ (76.6\%)\end{tabular} \\ \hline
\multicolumn{1}{|c||}{\begin{tabular}[c]{@{}c@{}}{\bf Deep-RBF}\\ 95\% - noiseless\end{tabular}} & \begin{tabular}[c]{@{}c@{}}3.22\%\\ —\end{tabular} & \begin{tabular}[c]{@{}c@{}}2.77\%\\ —\end{tabular} & \begin{tabular}[c]{@{}c@{}}100\%\\ (0\%)\end{tabular} & \begin{tabular}[c]{@{}c@{}}0.22\%\\ (3.77\%)\end{tabular} & \begin{tabular}[c]{@{}c@{}}100\%\\ (0\%)\end{tabular} & \begin{tabular}[c]{@{}c@{}}0.00\%\\ (9.77\%)\end{tabular} \\ \hline
\multicolumn{1}{|c||}{\begin{tabular}[c]{@{}c@{}}{\bf Deep-RBF}\\ 95\% - with noise\end{tabular}} & \begin{tabular}[c]{@{}c@{}}19.6\%\\ —\end{tabular} & \begin{tabular}[c]{@{}c@{}}16.5\%\\ —\end{tabular} & \begin{tabular}[c]{@{}c@{}}25.1\%\\ (10.2\%)\end{tabular} & \begin{tabular}[c]{@{}c@{}}100\%\\ (0\%)\end{tabular} & \begin{tabular}[c]{@{}c@{}}10.7\%\\ (47.3\%)\end{tabular} & \begin{tabular}[c]{@{}c@{}}100\%\\ (0\%)\end{tabular} \\ \hline
\end{tabular}
\end{small}
\end{center}
\end{table}

%%%%%%%%%%%%%%%%%%%%%%%%%
\section{Conclusion}

We have introduced a general formulation for the deep-RBF networks and proposed an example of cost functions for training these networks inspired from metric learning literature. We have also incorporated the reject option using a threshold on confidences. We have then examined two aspects of our method on the MNIST: first, we have evaluated its classification accuracy; second, we have assessed the vulnerability of the method to various adversarial attacks. We have observed that our method can attain significant classification accuracy on the MNIST test dataset and be very robust and resistant to adversarial examples at the same time using the reject option. By decreasing the threshold of rejection, the robustness of the method and the classification error rate both rise. This observation suggests that deep-RBF networks have the ability of rejection by the lack of data, especially when trained with our proposed cost function, and the noise injection regularization is applied. Other experiments involving negative examples could obtain further evidence for supporting this ability.

% has confirmed that the reject option can make deep-RBF networks robust due to the local nature of the RBF units at the last layer.
%In this subsection, we have shown experimentally that our methods are reasonably robust and resistant to adversarial attacks. It comes as no surprise that this fact is of great importance in practice. On the other hand, this fact also indicates that the proposed methods have the ability of rejection by lack of data, on which we insist. 

%%%%%%%%%%%%%%%%%%%%%%%%%%
%\section{Future works}
%\subsection{deep-RBF network with other network architectures on other datasets}
So far we have investigated our proposed method only on the MNIST dataset with a 2-layer ConvNet as the feature extractor. However, there exist many other datasets which are harder to classify, like ImageNet and CIFAR-10. In order to appraise the method better, we should compute the performance and robustness of our method on these datasets using more powerful network architectures as the feature extractor. Moreover, the proposed cost functions, i.e., ML and SoftML are just two samples of many possible metric learning losses. In the future, we should introduce and evaluate other metric learning cost functions for the deep-RBF networks.

\bibliographystyle{abbrvnat}
\bibliography{DeepRBF}
\end{document}